\documentclass{article}

\usepackage{microtype}
\usepackage{graphicx}
\usepackage{subfigure}
\usepackage[dvipsnames]{xcolor}
\usepackage{booktabs} % for professional tables

\usepackage{hyperref}

\usepackage[accepted]{icml2024}
\usepackage{amsmath}
\usepackage{amssymb}
\usepackage{mathtools}
\usepackage{amsthm}
\usepackage{colortbl}
\usepackage{adjustbox}
\usepackage{multirow}
\usepackage{multicol}
\usepackage{paralist}
\usepackage{caption}

\definecolor{codeblue}{rgb}{0.25,0.5,0.5}
\definecolor{codekw}{rgb}{0.85, 0.18, 0.50}
\definecolor{codegreen}{rgb}{0,0.6,0}
\definecolor{codegray}{rgb}{0.5,0.5,0.5}
\definecolor{codepurple}{rgb}{0.58,0,0.82}

\definecolor{coolblack}{rgb}{0.4, 0.01, 0.24}

\usepackage[capitalize,noabbrev]{cleveref}

\theoremstyle{plain}
\newtheorem{theorem}{Theorem}[section]

\theoremstyle{definition}
\newtheorem{definition}[theorem]{Definition}

\theoremstyle{remark}

\usepackage[textsize=tiny]{todonotes}

\definecolor{baselinecolor}{gray}{0.93}
\newcommand{\baseline}[1]{\cellcolor{baselinecolor}{#1}}

\icmltitlerunning{FiT: Flexible Vision Transformer for Diffusion Model}

\begin{document}

% \begin{document}

% \twocolumn[{
%     \renewcommand\twocolumn[1][]{#1}
%     \maketitle 
%     \input{figure/fig_teaser}
% }]
% \maketitle

\twocolumn[

\icmltitle{{\color{Maroon}FiT}: {\color{Maroon}F}lexible V{\color{Maroon}i}sion {\color{Maroon}T}ransformer for Diffusion Model}



\icmlsetsymbol{equal}{*}

% \begin{icmlauthorlist}
% \icmlauthor{Firstname1 Lastname1}{equal,yyy}
% \icmlauthor{Firstname2 Lastname2}{equal,yyy,comp}
% \icmlauthor{Firstname3 Lastname3}{comp}
% \end{icmlauthorlist}



\begin{icmlauthorlist}

\icmlsetsymbol{equal}{*}
\icmlauthor{Zeyu Lu}{pjlab,sjtu,equal}
\icmlauthor{Zidong Wang}{pjlab,thu,equal}
\icmlauthor{Di Huang}{pjlab,syu}
\icmlauthor{Chengyue Wu}{hku}
\icmlauthor{Xihui Liu}{hku}
\icmlauthor{Wanli Ouyang}{pjlab}
\icmlauthor{Lei Bai}{pjlab}
\end{icmlauthorlist}

\icmlaffiliation{sjtu}{Shanghai Jiao Tong University}
\icmlaffiliation{pjlab}{Shanghai Artificial Intelligence Laboratory}
\icmlaffiliation{thu}{Tsinghua University}
\icmlaffiliation{syu}{Sydney University}
\icmlaffiliation{hku}{The University of Hong Kong}

\icmlcorrespondingauthor{Lei Bai}{baisanshi@gmail.com}

\icmlkeywords{Machine Learning, ICML}

\vskip 0.3in

% \begin{figure*}[t]
%     \centering
%     \includegraphics[width=1.0\linewidth]{figure/teaser.pdf}
%     \caption{
%         performance
%     }
%     \label{fig:performance}
% \end{figure*}

\begin{center}
    \centering
    \vspace*{-0.3cm}
    \includegraphics[width=1.0\textwidth]{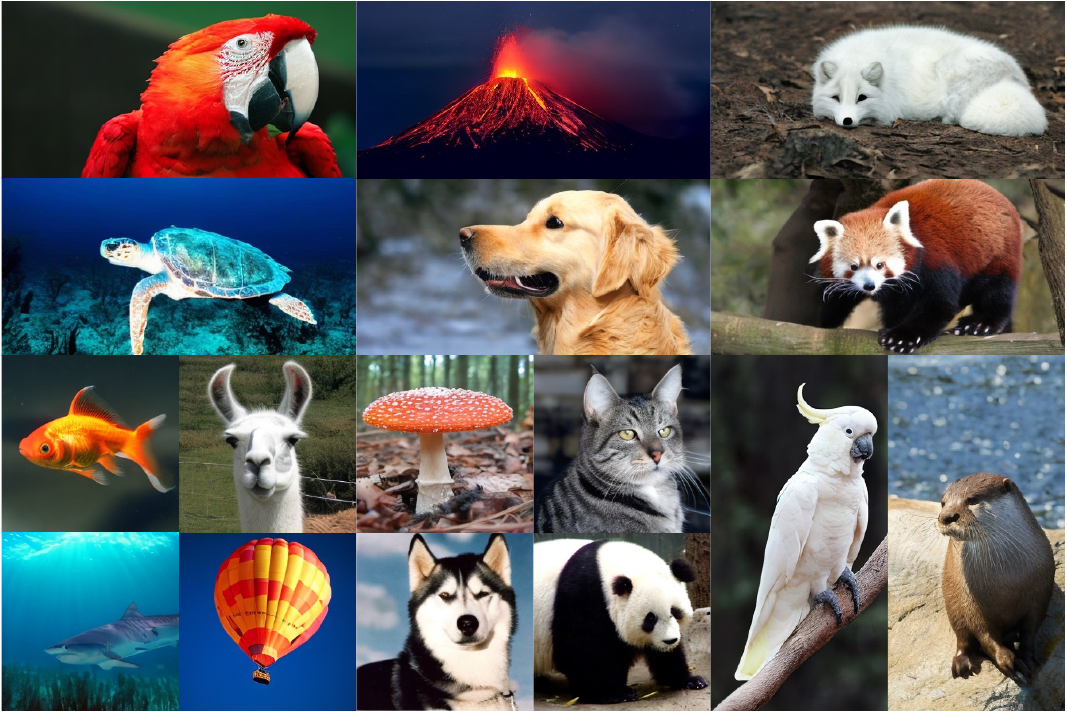}
    \vspace*{-0.6cm}
    \captionof{figure}{
    Selected samples from FiT-XL/2 models at resolutions of $256\times256$, $224\times448$ and $448\times224$.
    % \hd{FiT is capable of generating images at unrestricted resolutions and aspect ratios.}
    FiT is capable of generating images at unrestricted resolutions and aspect ratios.
    }
    \vspace*{-0.3cm}
    %    \textbf{ 
    %    Overview of our spatio-temporal Prediction Benchmark (PredBench)}. 
    %    % We comprehensively evaluate 4 crucial dimensions of 12 prevalent methods across 5 domains, covering 15 datasets. 
    %    % PredBench comprehensively evaluate 12 prevalent methods across 5 domains, covering 15 datasets, for the 4 crucial dimensions. 
    %    % Furthermore, our meticulous calibration of experiment settings, rigorous experiments, substantive analysis and enlightening insights will shed light on the future research of this field.
    %    PredBench conducts a comprehensive evaluation of 12 prevalent spatio-temporal prediction methods. It spans 5 distinct domains and covers 15 diverse datasets, thoroughly assessing the methods across 4 critical dimensions of prediction performance.
    %    % \textbf{Fig. 1. Overview of PredBench:} Our Spatio-temporal Prediction Benchmark (PredBench) conducts a comprehensive evaluation of 12 prevalent spatio-temporal prediction methods. It spans 5 distinct domains and covers 15 diverse datasets, thoroughly assessing the methods across 4 critical dimensions of prediction performance.
    % }
    % \caption{
    %     performance
    % }
    \label{fig:teaser}  
    \vspace*{+0.3cm} 
\end{center}

]



\printAffiliationsAndNotice{\icmlEqualContribution} % otherwise use the standard text.
% \printAffiliationsAndNotice{}

\begin{abstract}

\textit{Nature is infinitely resolution-free}.
In the context of this reality, existing diffusion models, such as Diffusion Transformers, often face challenges when processing image resolutions outside of their trained domain.
To overcome this limitation, we present the Flexible Vision Transformer (FiT), a transformer architecture specifically designed for generating images with \textit{unrestricted resolutions and aspect ratios}.
% Distinct from models that are constrained to fixed-resolution image training, 
% Distinct from fixed-resolution training, 
% FiT adopts a simple yet effective training strategy that accommodates varying aspect ratios, thus eliminating the need for image cropping.
% FiT adopts a simple yet effective training strategy that accommodates varying aspect ratios, thus enjoying the generalization ability in the resolution aspect and eliminating the basis from image cropping.
% FiT adopts a simple yet effective training strategy that accommodates varying aspect ratios for both training and inference. 
% This approach not only fosters resolution generalization but also eliminates biases introduced by image cropping. 
% \hd{
% % Distinct from viewing images as fixed-resolution grids, FiT model images as sequences of variable-length tokens. This novel modeling allows a simple yet effective training strategy that accommodates varying aspect ratios for both training and inference. 
% % This approach not only fosters resolution generalization but also eliminates biases introduced by image cropping. 
% Unlike traditional methods that perceive images as static-resolution grids, the FiT conceptualizes images as sequences of dynamically-sized tokens. This novel approach enables a flexible training strategy that effortlessly adapts to diverse aspect ratios during both training and inference phases, thus promoting resolution generalization and eliminating biases induced by image cropping.
% }
Unlike traditional methods that perceive images as static-resolution grids, FiT conceptualizes images as sequences of dynamically-sized tokens. This perspective enables a flexible training strategy that effortlessly adapts to diverse aspect ratios during both training and inference phases, thus promoting resolution generalization and eliminating biases induced by image cropping.
% Moving beyond the constraints of fixed-resolution image training, FiT employs a straightforward yet effective training strategy that not only accommodates various aspect ratios but also excels in resolution generalization. This approach effectively eliminates biases associated with image cropping and resizing.
% active resolution; resolution generalization;
% 1. cropping / resizing basis 
% 2. resolution generalization 
% Fit is able to remove fixed resolution 
Enhanced by a meticulously adjusted network structure and the integration of training-free extrapolation techniques, FiT exhibits remarkable flexibility in resolution extrapolation generation. 
Comprehensive experiments demonstrate the exceptional performance of FiT across a broad range of resolutions, showcasing its effectiveness both within and beyond its training resolution distribution.
% Repository available at \href{https://github.com/Inf-imagine/FiT}{https://github.com/Inf-imagine/FiT}.
Repository available at \href{https://github.com/whlzy/FiT}{https://github.com/whlzy/FiT}.
\end{abstract}
\section{Introduction}
% \xihui{Why start from transformer-based diffusion? Why not start from ``current t2i models face difficulties when generalizing to arbitrary resolutions''? The current version takes three paragraphs to introduce the challenge that we try to address.}

% Transformers are significantly impacting various domains within deep learning, as evidenced by the breakthroughs achieved with large language models (LLMs).
% In image generation, transformer-based generative models are beginning to outperform traditional UNet-based architectures, despite lacking inductive biases of UNet. These models demonstrate comparable effectiveness and remarkable scalability.

% The Diffusion Transformer (DiT), a novel class of diffusion models, underscores this progress.
% Originating from the latent diffusion model (LDM), 
% DiT and U-ViT integrate a vision transformer architecture for image generation.
% Its scalability underscores the potential benefits of integrating transformer architectures into diffusion models. Further developments, such as MDT and MaskDiT, which augment DiT with a decoder and masked modeling strategy, significantly enhance training efficiency and improve contextual and semantic learning in images.

% A critical challenge in deploying these models in practical applications is their performance across a wide range of image resolutions. For instance, as illustrated in Fig.~\ref{fig:imagenet}, images span various resolution spectrums. 
% Current text-to-image generation models face difficulties when generalizing to arbitrary resolutions. 
Current image generation models struggle with generalizing across arbitrary resolutions.
% However, current image generation models, including DiT, MDT, U-ViT, ADM and LDM, excel within specific resolution ranges but falter with images of varying resolutions. 
% For example, The Diffusion Transformer (DiT)~\cite{} family, excels within specific resolution ranges but falters with images of varying resolutions. 
The Diffusion Transformer (DiT) \cite{peebles2023scalable} family, while excelling within certain resolution ranges, falls short when dealing with images of varying resolutions.
% This limitation stems from the fact that DiT does not utilize dynamic resolution images during its training process, hindering its ability to adapt to different token lengths or resolutions effectively.
This limitation stems from the fact that DiT can not utilize dynamic resolution images during its training process, hindering its ability to adapt to different token lengths or resolutions effectively.

% To address these limitations, 
% To address the fixed resolution limitation, 
To overcome this limitation, 
we introduce the \textbf{Flexible Vision Transformer (FiT)}, which is adept at generating images at \textit{unrestricted resolutions and aspect ratios}. 
% FiT innovates through strategic enhancements in \textbf{network architecture}, \textbf{flexible training pipeline}, and \textbf{inference processes}.
% The key motivation lies in a brand new view of image data: Instead of modeling the image as fixed grid data, FiT models the image as variable-length token sequences. 
% By padding variable-length token sequences to a maximum length, FiT is capable of generating arbitrary tokens, enabling the unique ability of resolution-free generation. 
The key motivation is a novel perspective on image data modeling: 
% Rather than treating images as static grids of pixels, FiT interprets them as sequences of variable-length tokens. 
rather than treating images as static grids of fixed dimensions, FiT conceptualizes images as sequences of variable-length tokens.
This approach allows FiT to dynamically adjust the sequence length, thereby facilitating the generation of images at any desired resolution without being constrained by pre-defined dimensions. 
% By introducing padding to these variable-length token sequences up to a maximum defined length, FiT achieves the capability of producing images with unprecedented flexibility, marking a departure from resolution-dependent models to a resolution-agnostic paradigm
By efficiently managing variable-length token sequences and padding them to a maximum specified length, FiT unlocks the potential for resolution-independent image generation.
FiT represents this paradigm shift through significant advancements in \textbf{flexible training pipeline}, \textbf{network architecture}, and \textbf{inference processes}.

\begin{figure}[t]
    \centering
    \includegraphics[width=1\linewidth]{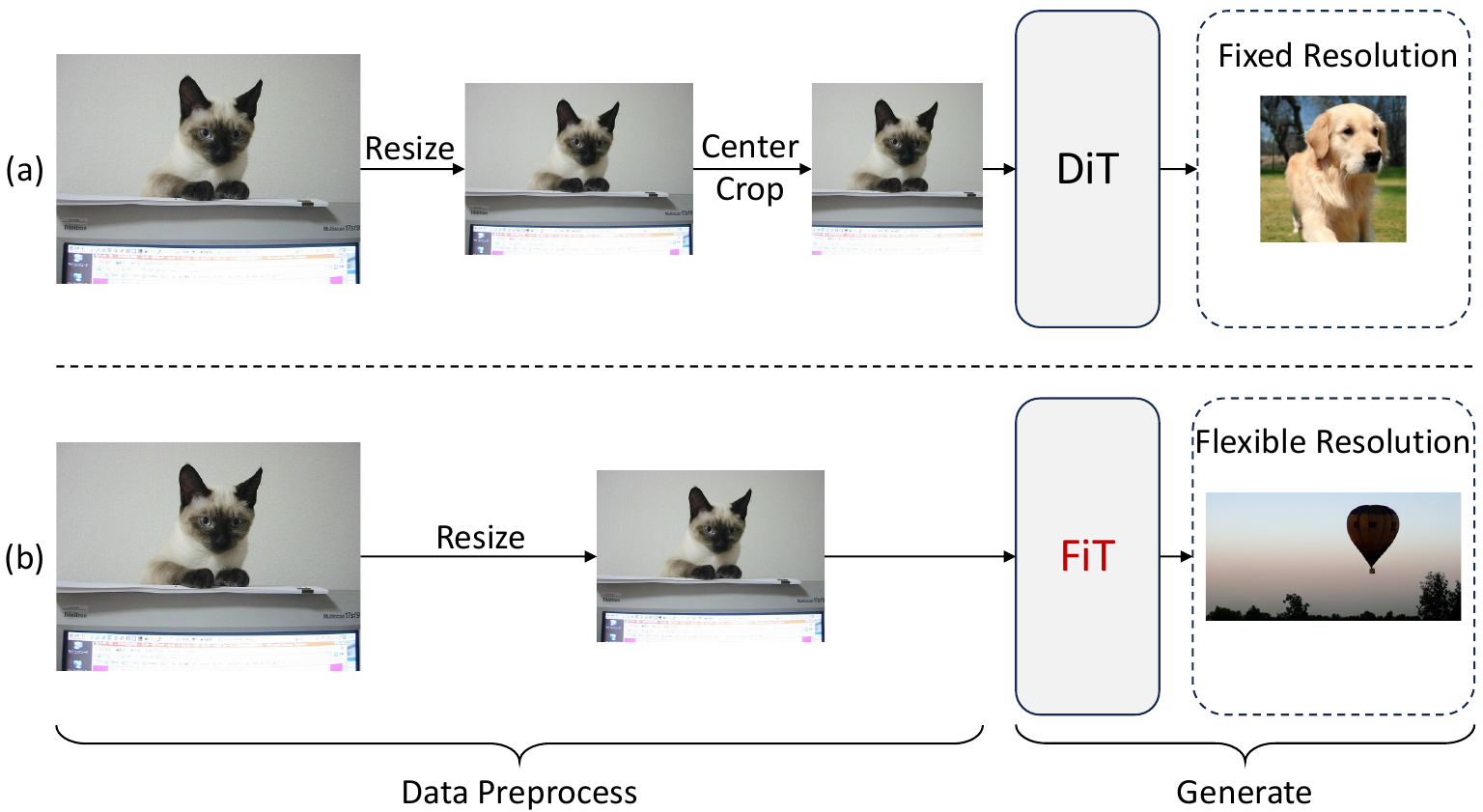}
    % \vspace{-0.3cm}
    \caption{
        Pipeline comparison between (a) DiT and (b) FiT.
    }
    % \vspace{-0.7cm}
    \label{fig:pipeline_overview}
\end{figure}

\noindent \textbf{Flexible Training Pipeline.} 
FiT uniquely preserves the original image aspect ratio during training, by viewing the image as a sequence of tokens. 
% by resizing higher-resolution images to fit within a maximum token limit, thus avoiding the cropping of higher-resolution images or resizing of lower-resolution images. 
% High-resolution images are resized to within a maximum token limit, avoiding the cropping of higher-resolution images or resizing of lower-resolution images. 
% This unique perspective allows FiT to adaptively resize high-resolution images to fit within a predefined maximum token limit, avoiding the cropping of higher-resolution images or resizing of lower-resolution images. 
This unique perspective allows FiT to adaptively resize high-resolution images to fit within a predefined maximum token limit, ensuring that no image, regardless of its original resolution, is cropped or disproportionately scaled.
This method ensures that the integrity of the image resolution is maintained, as shown in \cref{fig:pipeline_overview}, facilitating the ability to generate high-fidelity images at various resolutions.
To the best of our knowledge, FiT is the first transformer-based generation model to maintain diverse image resolutions throughout training.

% \hd{this figure is not that important. Instead, I think an overview figure for the main difference between FiT and other methods is better.}
\noindent \textbf{Network Architecture.} 
The FiT model evolves from the DiT architecture but addresses its limitations in resolution extrapolation.
% A critical component for generating images of various sizes is positional embedding. 
% Inspired by the success of Rotary Positional Embedding (RoPE) in large language models, we incorporate 2D RoPE instead of Absolute Positional Embedding.
% Furthermore, we replace DiT's Multilayer Perceptron (MLP) with Swish-Gated Linear Unit (SwiGLU), which is commonly used in recent large language models.
% % We replace Multi-Head Self-Attention (MHSA) in DiT with Masked Multi-Head Self-Attention (Masked MHSA) for dealing with the padding tokens used in our flexibel training pipeline.
% Moreover, we substitute DiT's Multi-Head Self-Attention (MHSA) with Masked Multi-Head Self-Attention (Masked MHSA) to handle the padding tokens introduced in our proposed flexible training pipeline.
% % Flexible training pipeline introduces padding tokens for flexibly packing dynamic sequences into a batch, so we 
One essential network architecture adjustment to handle diverse image sizes is the adoption of 2D Rotary Positional Embedding (RoPE)~\cite{su2024rope}, inspired by its success in enhancing large language models (LLMs) for length extrapolation~\cite{liu2023ropescaling}. We also introduce Swish-Gated Linear Unit (SwiGLU)~\cite{shazeer2020glu} in place of the traditional Multilayer Perceptron (MLP) and replace DiT’s Multi-Head Self-Attention (MHSA) with Masked MHSA to efficiently manage padding tokens within our flexible training pipeline.

% and replace DiT's Multilayer Perceptron (MLP) with Swish-Gated Linear Unit (SwiGLU), Multi-Head Self-Attention (MHSA) with Masked Multi-Head Self-Attention (Masked MHSA). 
% \xihui{Too specific. Any intuitions and motivations for replacing those layers?}

% \noindent \textbf{Flexible Training Pipeline.} 
% FiT uniquely preserves the original aspect ratio of images by resizing higher-resolution images to fit within a maximum token limit, thus avoiding the cropping of higher-resolution images or resizing of lower-resolution images.
% To the best of our knowledge, FiT is the first transformer-based generation model to maintain diverse image resolutions throughout training.

\noindent \textbf{Inference Process.} 
% Large language models use token length extrapolation techniques~\cite{peng2023yarn,ntkaware2023} to generate arbitrary lengths of texts. 
While large language models employ token length extrapolation techniques~\cite{peng2023yarn,ntkaware2023} for generating text of arbitrary lengths, a direct application of these technologies to FiT yields suboptimal results. We tailor these techniques for 2D RoPE, thereby enhancing FiT’s performance across a spectrum of resolutions and aspect ratios.

% \input{figure/fig_performance}

% Our FiT-XL/2 model trained only 1.7M steps outperforms all SOTA CNN models ("ADM-G,U", "LDM-4-G") and transformer models ("U-ViT-H/2-G, DiT-XL/2-G), surpassing it by a large margin on the resolutions of $160\times320$, $128\times384$, $320\times320$, $224\times448$, and $160\times480$.
% The performance of the FiT-XL/2 model significantly improved further after the adoption of our training-free extrapolation technique. compared to the DiT-XL, which was trained for 7M steps, the FiT-XL lags by 2 FID score points at a resolution of 256x256. However, at other resolutions, the FiT-XL d, significantly surpassing all other models.

% Our FiT-XL/2 model, after training for only 1.7 million steps, outperforms all state-of-the-art CNN ("ADM-G,U", "LDM-4-G") and transformer ("U-ViT-H/2-G, DiT-XL/2-G) models by a significant margin across resolutions of $160\times320$, $128\times384$, $320\times320$, $224\times448$, and $160\times480$. The performance of FiT-XL/2 significantly advances further with our training-free extrapolation technique. Compared to the DiT-XL, which underwent training for 7 million steps, FiT-XL lags by merely 2 FID score points at a resolution of 256x256 but significantly surpasses all competing models at other resolutions.
Our highest Gflop FiT-XL/2 model, after training for only 1.8 million steps on \textit{ImageNet-256}~\cite{deng2009imagenet} dataset, outperforms all state-of-the-art CNN and transformer models by a significant margin across resolutions of $160\times320$, $128\times384$, $320\times320$, $224\times448$, and $160\times480$. The performance of FiT-XL/2 significantly advances further with our training-free resolution extrapolation method. Compared to the baseline DiT-XL/2 training for 7 million steps, FiT-XL/2 lags slightly at the resolution of $256\times256$ but significantly surpasses it at all other resolutions.
% In summary, our contributions are threefold:
% \begin{compactitem}
%     \item Introduction of FiT, a scalable and flexible vision transformer tailored for image diffusion models, capable of generating images at any resolution and aspect ratio.
    
%     \item Implementation of innovative design features in FiT, including a unique transformer architecture for dynamic token length, a flexible training pipeline that eliminates the need for cropping, and an inference strategy optimized for arbitrary resolution generation.
    
%     \item Demonstration that the FiT-XL/2 model achieves state-of-the-art performance across a variety of resolution settings.
% \end{compactitem}
In summary, our contributions lie in the novel introduction of FiT, a flexible vision transformer tailored for diffusion models, capable of generating images at any resolution and aspect ratio. 
We present three innovative design features in FiT, including 
a flexible training pipeline that eliminates the need for cropping, 
a unique transformer architecture for dynamic token length modeling, 
and a training-free resolution extrapolation method for arbitrary resolution generation.
% We present three key innovations: a novel transformer architecture capable of dynamic token length adaptation, a flexible training pipeline that obviates the need for image cropping, and an optimized inference strategy for arbitrary resolution generation. 
Strict experiments demonstrate that the FiT-XL/2 model achieves state-of-the-art performance across a variety of resolution and aspect ratio settings. 

% \xihui{It's not common to write the contribution in this way. usually three different keypoints, or two keypoints + experiment highlight. If keep this logic, you can merge the three items into a single paragraph.}

% \input{figure/fig_imagenet}
\section{Related Work}
% \hd{Delete in half.}

\noindent\textbf{Diffusion Models.}
Denoising Diffusion Probabilistic Models (DDPMs)~\cite{ho2020denoising,saharia2022photorealistic,radford2021learning} and score-based models~\cite{hyvarinen2005estimation,song2020score} have exhibited remarkable progress in the context of image generation tasks. The Denoising Diffusion Implicit Model (DDIM)~\citet{song2020denoising}, offers An accelerated sampling procedure. Latent Diffusion Models (LDMs)~\cite{rombach2022high} establishes a new benchmark of training deep generative models to reverse a noise process in the latent space, through the use of VQ-VAE~\cite{esser2021taming}.

% Denoising Diffusion Probabilistic Models (DDPMs) signify a considerable advancement in the field of diffusion models~\cite{ho2020denoising,saharia2022photorealistic,radford2021learning}. Before their development, Generative Adversarial Networks (GANs) were predominantly favored for generative tasks~\cite{goodfellow2014generative}. However, diffusion and score-based generative models have since exhibited remarkable progress, especially in the context of image generation tasks~\cite{hyvarinen2005estimation,song2020score}. An accelerated sampling procedure, the Denoising Diffusion Implicit Model (DDIM), was proposed by ~\citet{song2020denoising}, offering a more efficient approach. The concept of latent space modeling, a critical technique in deep generative models, which utilized encoder-decoder Variational Autoencoder (VAE)\cite{kingma2013auto} architectures for reconstruction purposes, resulting in the creation of Latent Diffusion Models (LDMs)~\cite{rombach2022high}. Upon their debut, LDMs established a new benchmark for sample quality by training deep generative models to reverse a noise corruption process in the latent space.

\noindent\textbf{Transformer Models.}
% The Transformer model~\cite{vaswani2017attention}, has successfully supplanted domain-specific architectures in a variety of fields including, but not limited to, language~\cite{brown2020language}, vision~\cite{dosovitskiy2020image}, and multi-modality~\cite{team2023gemini}. Its exceptional scalability concerning model size, computational training, and data has been well-demonstrated~\cite{hoffmann2022training,kaplan2020scaling}, making it a robust choice for generic auto-regressive models. Beyond the realm of language, Transformers have been effectively deployed for pixel prediction~\cite{chen2020generative} and auto-regressive prediction of discrete codebooks~\cite{esser2021taming,yu2022scaling,chang2023muse}, achieving impressive scaling up to 20 billion parameters. Additionally, Transformers have been instrumental in the development of denoising diffusion probabilistic models~\cite{ho2020denoising}, facilitating the synthesis of non-spatial data such as Contrastive Language–Image Pretraining (CLIP) image embeddings~\cite{saharia2022photorealistic,radford2021learning}.
The Transformer model~\cite{vaswani2017attention}, has successfully supplanted domain-specific architectures in a variety of fields including, but not limited to, language~\cite{brown2020language,chowdhery2023palm}, vision~\cite{dosovitskiy2020image}, and multi-modality~\cite{team2023gemini}. 
In vision perception research, most efforts~\cite{touvron2019fixing,touvron2021training,liu2021swin,liu2022convnet} that focus on resolution are aimed at accelerating pretraining using a fixed, low resolution. On the other hand, NaViT~\cite{dehghani2023patch} implements the 'Patch n’ Pack' technique to train ViT using images at their natural, 'native' resolution.
Notably, transformers have been also explored in the denoising diffusion probabilistic models~\cite{ho2020denoising} to synthesize images. DiT~\cite{peebles2023scalable} is the seminal work that utilizes a vision transformer as the backbone of LDMs and can serve as a strong baseline. Based on DiT architecture, MDT~\cite{gao2023masked} introduces a masked latent modeling approach, which requires two forward-runs in training and inference. U-ViT~\cite{bao2023all} treats all inputs as tokens and incorporates U-Net architectures into the ViT backbone of LDMs. DiffiT~\cite{hatamizadeh2023diffit} introduces a time-dependent self-attention module into the DiT backbone to adapt to different stages of the diffusion process. We follow the LDM paradigm of the above methods and further propose a novel flexible image synthesis pipeline.

\noindent \textbf{Length Extrapolation in LLMs.}
RoPE (Rotary Position Embedding)~\cite{su2024rope} is a novel positional embedding that incorporates relative position information into absolute positional embedding. It has recently become the dominant positional embedding in a wide range of LLM (Large Language Model) designs~\cite{chowdhery2024palm,touvron2023llama,touvron2023llama2}. Although RoPE enjoys valuable properties, such as the flexibility of sequence length, its performance drops when the input sequence surpasses the training length. Many approaches have been proposed to solve this issue. PI (Position Interpolation)~\cite{chen2023pi} linearly down-scales the input position indices to match the original context window size, while NTK-aware~\cite{ntkaware2023} changes the rotary base of RoPE. YaRN (Yet another RoPE extensioN)~\cite{peng2023yarn} is an improved method to efficiently extend the context window. RandomPE~\cite{ruoss2023randompe} sub-samples an ordered set of positions from a much larger range of positions than originally observed in training or inference. xPos~\cite{sun2022xpos} incorporates long-term decay into RoPE and uses blockwise causal attention for better extrapolation performance. Our work delves deeply into the implementation of RoPE in vision generation and on-the-fly resolution extrapolation methods.

\section{Flexible Vision Transformer for Diffusion}

\begin{figure*}[t]
    \centering
    \includegraphics[width=1.0\linewidth]{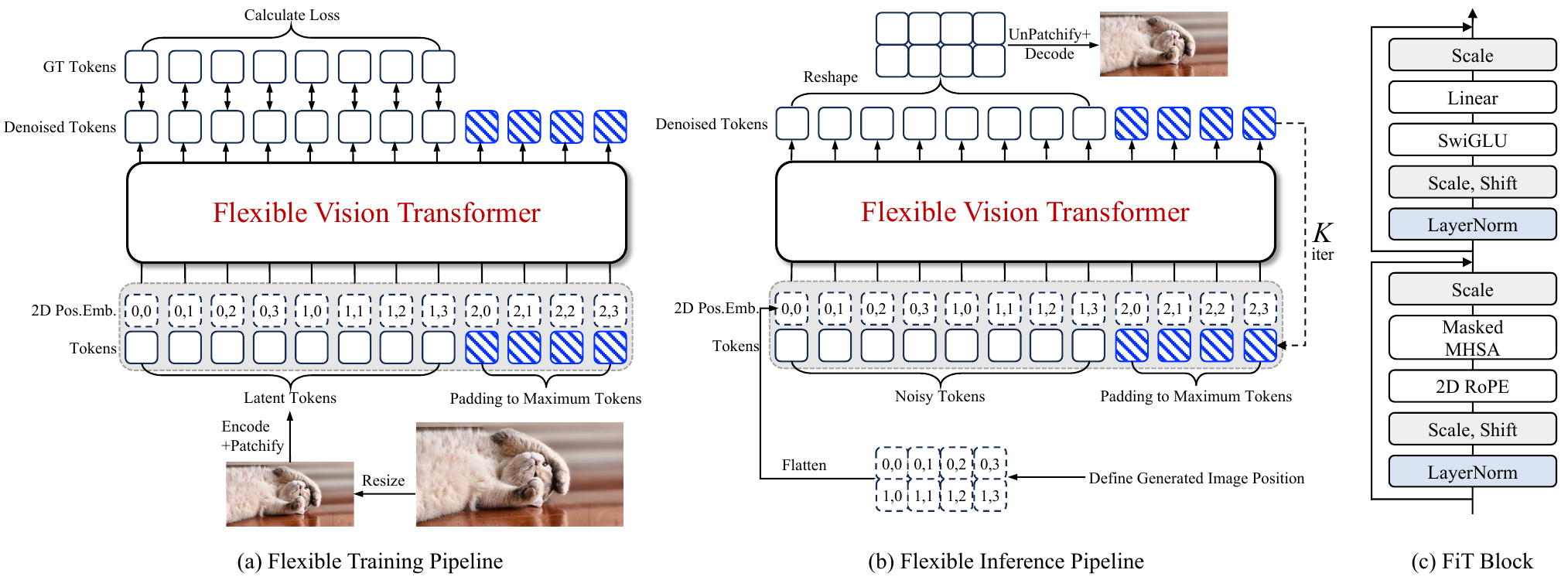}
    % \vspace{-0.2cm}
    \caption{
        % \hd{Not Good. Remake.}
        Overview of (a) flexible training pipeline, (b) flexible inference pipeline, and (c) FiT block.
    }
    % \vspace{-0.3cm}
    \vskip -0.15 in
    \label{fig:overview}
\end{figure*}

\subsection{Preliminary}
% \hd{We should consider deleting this section.}
\label{subsec:method_preliminary}

% \hd{You need to clarify: why do you mention RoPE, NTK-aware Interpolation, and YaRN here?}

\noindent \textbf{1-D RoPE (Rotary Positional Embedding)} \cite{su2024rope} is a type of position embedding that unifies absolute and relative PE, exhibiting a certain degree of extrapolation capability in LLMs. Given the m-th key and n-th query vector as $\mathbf{q}_m, \mathbf{k}_n \in \mathbb{R}^{|D|}$, 1-D RoPE multiplies the bias to the key or query vector in the complex vector space:
\begin{equation}
    f_q(\mathbf{q}_m, m) = e^{im\Theta} \mathbf{q}_m, \quad
    f_k(\mathbf{k}_n, n) = e^{in\Theta} \mathbf{k}_n
    \label{eq:1d_rope}
\end{equation}
where $\Theta=\mathrm{Diag}(\theta_1, \cdots, \theta_{|D|/2})$ is rotary frequency matrix with $\theta_d = b^{-2d/|D|}$ and rotary base $b=10000$. In real space, given $l=|D|/2$, the rotary matrix $e^{im\Theta}$ equals to:
\begin{equation}
% \begin{aligned}
    % &f_q(\mathbf{q}_m, m) = \mathrm{Re}[e^{im\theta} \mathbf{q}_m] = \\ 
    \begin{bmatrix}
        \cos m\theta_1 & -\sin m\theta_1 & \cdots & 0 & 0 \\
        \sin m\theta_1 & \cos m\theta_1 & \cdots & 0 & 0 \\
        \vdots & \vdots & \ddots & \vdots & \vdots \\
        0 & 0 & \cdots & \cos m\theta_l & -\sin m\theta_l \\
        0 & 0 & \cdots & \sin m \theta_l & \cos m\theta_l
    \end{bmatrix}
    \label{eq:real_space_rotary_matrix}
% \end{aligned}
\end{equation}

The attention score with 1-D RoPE is calculated as:
\begin{equation}
    A_n = \mathrm{Re}\langle f_q(\mathbf{q}_m, m), f_k(\mathbf{k}_n, n)\rangle
    \label{eq:1d_rope_attn}
\end{equation}

\noindent \textbf{NTK-aware Interpolation} \cite{ntkaware2023} is a training-free length extrapolation technique in LLMs. To handle the larger context length $L_{\text{test}}$ than maximum training length $L_{\text{train}}$, it modifies the rotary base of 1-D RoPE as follows:
\begin{equation}
    b'=b\cdot s^{\frac{|D|}{|D|-2}}, 
    \label{eq:1d_ntk}
\end{equation}
where the scale factor $s$ is defined as: 
\begin{equation}
    s=\max(\frac{L_{test}}{L_{train}}, 1.0).
    \label{eq:1d_rope_scale_factor}
\end{equation}

% PartNTK在YaRN的论文中叫NTK-by-parts，但是在YaRN的代码中叫PartNTK
% YaRN论文种对于PartNTK的定义有误，此处不再采用
% $r(d)=\frac{L_{\text{train}}}{2\pi b^{2d/|D|}}$
\noindent \textbf{YaRN (Yet another RoPE extensioN) Interpolation} \cite{peng2023yarn} introduces the ratio of dimension $d$ as $r(d)={L_{\text{train}}}/({2\pi b^{2d/|D|}})$, and modifies the rotary frequency as:
\begin{equation}
    \theta'_d=\left(1-\gamma(r(d))\right)\frac{\theta_d}{s}+\gamma(r(d))\theta_d,
    \label{eq:1d_partntk}
\end{equation}

where $s$ is the aforementioned scale factor, and $\gamma(r(d))$ is a ramp function with  extra hyper-parameters $\alpha, \beta$:
\begin{equation}
    \gamma(r)=\begin{cases}
    0,&\text{if }r<\alpha\\
    1,&\text{if }r>\beta\\
    \frac{r-\alpha}{\beta-\alpha},&\text{otherwise}.
    \end{cases}
    \label{eq:1d_partntk_ramp}
\end{equation}

Besides, it incorporates a 1D-RoPE scaling term as:
\begin{equation}
    f'_q(\mathbf{q}_m, m) = \frac{1}{\sqrt{t}}f_q(\mathbf{q}_m, m), 
    f'_k(\mathbf{k}_n, n) = \frac{1}{\sqrt{t}} f_k(\mathbf{k}_n, n),
    \label{eq:1d_yarn_scale}
\end{equation}
where $\frac{1}{\sqrt{t}}=0.1\ln(s)+1$.

\subsection{Flexible Training and Inference Pipeline}

% 过去的DiT在固定分辨率的图像上进行训练， 
% 
% 我们首先设置最大token长度 L
% 
%  flexible training startegy

% training和inference都要写清楚，比如encode是怎么encode，resize的做法，patchify前后的维度，结构上差异在于把DiT的backbone改为了FiT的backbone。。。。。。

% Deep neural networks are commonly trained and executed using batches of input data. To enable efficient processing on current hardware, this requires a fixed batch shape, which in turn means a consistent image size for computer vision applications. 
Modern deep learning models, constrained by the characteristics of GPU hardware, are required to pack data into batches of uniform shape for parallel processing.
Due to the diversity in image resolutions, as shown in Fig.~\ref{fig:imagenet}, DiT resizes and crops the images to a fixed resolution $256\times256$.
While resizing and cropping as a means of data augmentation is a common practice, this approach introduces certain biases into the input data. 
These biases will directly affect the final images generated by the model, including blurring effects from the transition from low to high resolution and information lost due to the cropping (more failure samples can be found in Appendix ~\ref{appendix_sample}). %\zeyu{more analysis}
% Due to the diversity in image resolutions, it's a standard practice in computational vision to resize and crop images from various resolutions to a standard resolution. For generative models, while random cropping is a natural form of data augmentation, it can unintentionally appear in the generated samples. Consequently, the model may inadvertently learn and replicate these blurring and cropping effects.

To this end, we propose a flexible training and inference pipeline, as shown in Fig.~\ref{fig:overview} (a, b). 
\textit{In the preprocessing phase}, we avoid cropping images or resizing low-resolution images to a higher resolution. Instead, we only resize high-resolution images to a predetermined maximum resolution limit, $HW\leqslant256^2$.
\textit{In the training phase}, FiT first encodes an image into latent codes with a pre-trained VAE encoder.
By patchfiying latent codes to latent tokens, we can get sequences with different lengths $L$.
To pack these sequences into a batch, we pad all these sequences to the maximum token length $L_{max}$ using padding tokens.
Here we set $L_{max}=256$ to match the fixed token length of DiT. 
The same as the latent tokens, we also pad the positional embeddings to the maximum length for packing.
Finally, we calculate the loss function only for the denoised output tokens, while discarding all other padding tokens.
\textit{In the inference phase}, we firstly define the position map of the generated image and sample noisy tokens from the Gaussian distribution as input. After completing $K$ iterations of the denoising process, we reshape and unpatchfiy the denoised tokens according to the predefined position map to get the final generated image.

% In the following parts, we will introduce the two key components, (1) FiT architecture, and (2) resolution extrapolation technique of FiT.

\subsection{Flexible Vision Transformer Architecture}
% \hd{Re-write.}

% Based on the flexible training strategy, we aim to find the suitable architecture which can stablely training in different resolutions.
Building upon the flexible training pipeline, our goal is to find an architecture that can stably train across various resolutions and generate images with arbitrary resolutions and aspect ratios, as shown in \cref{fig:overview} (c).
Motivated by some significant architectural advances in LLMs, we conduct a series of experiments to explore architectural modifications based on DiT, see details in \cref{subsec:exp_fit_architecture}.

\noindent\textbf{Replacing MHSA with Masked MHSA.}
The flexible training pipeline introduces padding tokens for flexibly packing dynamic sequences into a batch. 
During the forward phase of the transformer, it is crucial to facilitate interactions among noised tokens while preventing any interaction between noised tokens and padding tokens. 
The Multi-Head Self-Attention (MHSA) mechanism of original DiT is incapable of distinguishing between noised tokens and padding tokens. 
To this end, we use Masked MHSA to replace the standard MHSA. 
We utilize the sequence mask $M$ for Masked Attention, where noised tokens are assigned the value of $0$, and padding tokens are assigned the value of negative infinity (\textit{-inf}), which is defined as follows:

\begin{equation}
\text{Masked Attn.}(Q_i, K_i, V_i) = \text{Softmax}\left(\frac{Q_iK_i^T}{\sqrt{d_k}}+M\right)V_i 
\end{equation}
% \begin{equation}
% M=\left\{ 
%     \begin{aligned}
%     &1 \\
%     &-\inf \\
%     \end{aligned}
% \right.
% \end{equation}

where $Q_i$, $K_i$, $V_i$ are the query, key, and value matrices for the $i$-th head.

\noindent\textbf{Replacing Absolute PE with 2D RoPE.}
% 用另一种表述来写这一部分
% Rotary Positional Embedding (RoPE) is a type of position embedding that unifies absolute and relative PE, exhibiting a degree of extrapolation capability in LLMs in comparison to Absolute PE.
% We aspire for our architecture to not only generate images with resolution within the training distribution range, but also those with resolution beyond the training distribution range.
% To this end, we use 2D RoPE for positional information injection and the self-attention score  $a$ is defined as follows:
% \begin{equation}
% \begin{aligned}
% % Q_i K_i^T 
% &a((x_0,y_0), (x_1,y_1)) = \mathrm{Re} \langle f(q,(x_0,y_0)), f(k,(x_1,y_1)) \rangle \\
% &= \mathrm{Re} 
% \left[ 
% \sum^{d/4-1}_{j=0} (q_{2j}+iq_{2j+1})(k_{2j}-ik_{2j+1})e^{i(x_0-x_1)\theta_j} + \right. \\
% &\left. \sum^{d/4-1}_{j=0} (q_{2j}+iq_{2j+1})(k_{2j}-ik_{2j+1})e^{i(y_0-y_1)\theta_j} 
% \right] \\
% &= \sum^{d/4-1}_{j=0} [ (q_{2j}k_{2j} + q_{2j+1}k_{2j+1})\cos((x_0-x_1)\theta_j) + \\ 
% &(q_{2j}k_{2j+1} - q_{2j+1}k_{2j})\sin((x_0-x_1)\theta_j)]+ \\
% &\sum^{d/4-1}_{j=0} [ (q_{2j}k_{2j} + q_{2j+1}k_{2j+1})\cos((y_0-y_1)\theta_j) + \\
% &(q_{2j}k_{2j+1} - q_{2j+1}k_{2j})\sin((y_0-y_1)\theta_j)]\\
% \end{aligned} 
% \label{2drope}
% \end{equation}
% where $(x_0,y_0)$, $(x_1,y_1)$ are the position indices of the embedding vectors $q$ and $k$, the subscripts of $q$ and $k$ denote the dimensions of the attention head, $\theta^n=10000^{-2n/d}$ is the rotary base.
We observe that vision transformer models with absolute positional embedding fail to generalize well on images beyond the training resolution, as in \cref{subsec:exp_extrapolation_design,subsec:exp_out_of_distribution}. Inspired by the success of 1D-RoPE in LLMs for length extrapolation~\cite{liu2023ropescaling}, we utilize 2D-RoPE to facilitate the resolution generalization in vision transformer models. Formally, we calculate the 1-D RoPE for the coordinates of height and width separately. Then such two 1-D RoPEs are concatenated in the last dimension.  Given 2-D coordinates of width and height as $\{(w, h) \Big| 1\leqslant w \leqslant W, 1 \leqslant h \leqslant H\}$, the 2-D RoPE is defined as:
\begin{equation}
\begin{aligned}
    & f_q(\mathbf{q}_m, h_m, w_m) = [e^{i h_m \Theta} \mathbf{q}_m \parallel e^{i w_m \Theta} \mathbf{q}_m ], \\
    & f_k(\mathbf{k}_n, h_n, w_n) = [e^{i h_n \Theta} \mathbf{k}_n \parallel e^{i w_n \Theta} \mathbf{k}_n ], \\
\end{aligned}
\label{eq:2d_rope}
\end{equation}
where $\Theta=\mathrm{Diag}(\theta_1, \cdots, \theta_{|D|/4})$, and $\parallel$ denotes concatenate two vectors in the last dimension. Note that we divide the $|D|$-dimension space into $|D|/4$-dimension subspace to ensure the consistency of dimension, which differs from $|D|/2$-dimension subspace in 1-D RoPE. Analogously, the attention score with 2-D RoPE is:
\begin{equation}
    A_n = \mathrm{Re}\langle f_q(\mathbf{q}_m, h_m, w_m), f_k(\mathbf{k}_n, h_n, w_n)\rangle.
    \label{eq:2d_rope_attn}
\end{equation}

It is noteworthy that there is no cross-term between $h$ and $w$ in 2D-RoPE and attention score $A_n$, so we can further decouple the rotary frequency as $\Theta_h$ and $\Theta_w$, resulting in the decoupled 2D-RoPE, which will be discussed in \cref{subsec:method_extrapolation} and more details can be found in Appendix ~\ref{detail_attn_score}.

% Formally, we decouple the positional indices of the width and height of an image to calculate 2-D RoPE. Given an index of width and height as $x_m (1\leqslant x_m \leqslant W)$ and $y_n (1 \leqslant y_n \leqslant H)$, the 2-D RoPE is defined as:
% \begin{equation}
% \begin{aligned}
%     & f_q(\mathbf{q}_m, x_m, y_m) = [e^{i x_m \theta} \mathbf{q}_m \parallel e^{i y_m \theta} \mathbf{q}_m ], \\
%     & f_k(\mathbf{k}_n, x_n, y_n) = [e^{i x_n \theta} \mathbf{k}_n \parallel e^{i y_n \theta} \mathbf{k}_n ], \\
% \end{aligned}
% \end{equation}
% where $\theta=\mathrm{Diag}(\theta_1, \cdots, \theta_{|D|/4})$, and $\parallel$ denotes concatenate two vectors in the last dimension. Analogously, the attention score with 2-D RoPE is:
% \begin{equation}
%     A_n = \mathrm{Re}\langle f_q(\mathbf{q}_m, x_m, y_m), f_k(\mathbf{k}_n, x_n, y_n)\rangle
% \end{equation}

\noindent\textbf{Replacing MLP with SwiGLU.} We follow recent LLMs like LLaMA~\cite{touvron2023llama,touvron2023llama2}, and replace the MLP in FFN with SwiGLU, which is defined as follows:
\begin{equation}
\begin{split}
\text{SwiGLU}(x, W, V) = \text{SiLU}(x W) \otimes (x V) \\
\text{FFN}(x) = \text{SwiGLU}(x, W_1, W_2)W_3
\end{split}
\end{equation}

% element-wise multiplication == Hadmard Product
where $\otimes$ denotes Hadmard Product, $W_1$, $W_2$, and $W_3$ are the weight matrices without bias, $\text{SiLU}(x) = x \otimes \sigma(x)$. Here we will use SwiGLU as our choice in each FFN block.

\subsection{Training Free Resolution Extrapolation}
\label{subsec:method_extrapolation}

We denote the inference resolution as ($H_{\text{test}}$, $W_{\text{test}}$).  Our FiT can handle various resolutions and aspect ratios during training, so we denote training resolution as $L_{\text{train}} = \sqrt{L_{\text{max}}}$.

% $s=\max(\frac{\max(H_{\text{test}}, W_{\text{test}})}{L_{train}}, 1.0)$
By changing the scale factor in \cref{eq:1d_rope_scale_factor} to $s=\max( {\max(H_{\text{test}}, W_{\text{test}})} / {L_{train}}, 1.0)$, we can directly implement the positional interpolation methods in large language model extrapolation on 2D-RoPE, which we call vanilla NTK and YaRN implementation. Furthermore, we propose vision RoPE interpolation methods by using the decoupling attribute in decoupled 2D-RoPE. We modify \cref{eq:2d_rope} to: 
\begin{equation}
\begin{aligned}
    & \hat{f}_q(\mathbf{q}_m, h_m, w_m) = [e^{i h_m \Theta_h} \mathbf{q}_m \parallel e^{i w_m \Theta_w} \mathbf{q}_m ], \\
    & \hat{f}_k(\mathbf{k}_n, h_n, w_n) = [e^{i h_n \Theta_h} \mathbf{k}_n \parallel e^{i w_n \Theta_w} \mathbf{k}_n ], \\
\end{aligned}
\label{eq:2d_rope_decoupled}
\end{equation}
where $\Theta_h=\{\theta^h_d=b_h^{-2d/|D|}, 1\leqslant d\leqslant \frac{|D|}{2}\}$ and $\Theta_w=\{\theta^w_d=b_w^{-2d/|D|}, 1\leqslant d\leqslant \frac{|D|}{2}\}$ are calculated separately. Accordingly, the scale factor of height and width is defined separately as 
\begin{equation}
    s_h = \max(\frac{H_{\text{test}}}{L_{\text{train}}}, 1.0), \quad s_w = \max(\frac{W_{\text{test}}}{L_{\text{train}}}, 1.0).
\end{equation}

\begin{definition}
    \textit{The Definition of VisionNTK Interpolation is a modification of NTK-aware Interpolation by using \cref{eq:2d_rope_decoupled} with the following rotary base.} 
    \begin{equation}
        b_h = b \cdot s_h^{\frac{|D|}{|D|-2}}, \quad
        b_w = b \cdot s_w^{\frac{|D|}{|D|-2}},
    \label{eq:2d_rope_visntk}
    \end{equation}
    where $b=10000$ is the same with \cref{eq:1d_rope} 
    \label{def:vision_ntk}
\end{definition}

\begin{definition}
    \textit{The Definition of VisionYaRN Interpolation is a modification of YaRN Interpolation by using \cref{eq:2d_rope_decoupled} with the following rotary frequency.}
    \begin{equation}
    \begin{aligned}
        & \theta^h_d = (1-\gamma(r(d))\frac{\theta_d}{s_h} + \gamma(r(d))\theta_d, \\
        & \theta^w_d = (1-\gamma(r(d))\frac{\theta_d}{s_w} + \gamma(r(d))\theta_d, \\    
    \end{aligned}
    \end{equation}
    where $\gamma(r(d))$ is the same with \cref{eq:1d_partntk}.
    \label{def:vision_yarn}
\end{definition}

It is worth noting that VisionNTK and VisionYaRN are training-free positional embedding interpolation approaches, used to alleviate the problem of position embedding out of distribution in extrapolation. When the aspect ratio equals one, they are equivalent to the vanilla implementation of NTK and YaRN. They are especially effective in generating images with arbitrary aspect ratios, see \cref{subsec:exp_extrapolation_design}.

% \begin{definition}
% 用了式9那就要说明一下式10的attention score x和y没有交叉相，所以才能分别考虑
% \textit{(Definition of Vision RoPE) For 2D RoPE-based vision transformer pre-trained with rotary base $\theta$. We can define the two rotary base $\theta^x$, $\theta^y$ for 2d coordinate and the RoPE Eqn.~\ref{2drope} can be reformulated as follows:}

% \begin{equation}
% \begin{aligned}
% &a((x_0,y_0), (x_1,y_1)) = \\
% &\sum^{d/4-1}_{j=0} [ (q_{2j}k_{2j} + q_{2j+1}k_{2j+1})\cos((x_0-x_1)\theta^x_j) + \\ 
% &(q_{2j}k_{2j+1} - q_{2j+1}k_{2j})\sin((x_0-x_1)\theta^x_j)]+ \\
% &\sum^{d/4-1}_{j=0} [ (q_{2j}k_{2j} + q_{2j+1}k_{2j+1})\cos((y_0-y_1)\theta^y_j) + \\
% &(q_{2j}k_{2j+1} - q_{2j+1}k_{2j})\sin((y_0-y_1)\theta^y_j)]\\
% \end{aligned} 
% \end{equation}
% \end{definition}
\section{Experiments}
\begin{table*}[t]
\centering
% \addtolength{\tabcolsep}{-1pt}

\begin{adjustbox}{max width=1.\textwidth}
\begin{tabular}{l|c|ccccc|ccccc|ccccc}
\toprule[1.2pt]
\multirow{2}*{Method}&\multirow{2}*{Train Cost}&\multicolumn{5}{c|}{256$\times$256 (1:1)}&\multicolumn{5}{c|}{160$\times$320 (1:2)}&\multicolumn{5}{c}{128$\times$384 (1:3)} \\

& 

& \textbf{FID$\downarrow$} & \textbf{sFID$\downarrow$} & \textbf{IS$\uparrow$} & \textbf{Prec.$\uparrow$} & \textbf{Rec.$\uparrow$}
& \textbf{FID$\downarrow$} & \textbf{sFID$\downarrow$} & \textbf{IS$\uparrow$} & \textbf{Prec.$\uparrow$} & \textbf{Rec.$\uparrow$}  
& \textbf{FID$\downarrow$} & \textbf{sFID$\downarrow$} & \textbf{IS$\uparrow$} & \textbf{Prec.$\uparrow$} & \textbf{Rec.$\uparrow$}  \\
% & \textbf{FID} & \textbf{sFID} & \textbf{IS} & \textbf{Prec.} & \textbf{Rec.}
% & \textbf{FID} & \textbf{sFID} & \textbf{IS} & \textbf{Prec.} & \textbf{Rec.}  
% & \textbf{FID} & \textbf{sFID} & \textbf{IS} & \textbf{Prec.} & \textbf{Rec.}  \\

\midrule
BigGAN-deep & - & 6.95 & 7.36 & 171.4 & 0.87 & 0.28 &-&-&-&-&-&-&-&-&-&- \\
StyleGAN-XL & - & 2.30 & 4.02 & 265.12 & 0.78 & 0.53 &-&-&-&-&-&-&-&-&-&- \\
MaskGIT & 1387k$\times$256 &6.18 &- &182.1 &0.80 &0.51 &-&-&-&-&-&-&-&-&-&- \\
CDM & - & 4.88 & - & 158.71 & - & - &-&-&-&-&-&-&-&-&-&- \\

\midrule

U-ViT-H/2-G (cfg=1.4) & 500k$\times$1024 & 2.35 &5.68&265.02&0.82&0.57&6.93&12.64&175.08&0.67&0.63&196.84&95.90& 7.54& 0.06&0.27\\

ADM-G,U & 1980k$\times$256 &3.94 &6.14 &215.84 &0.83 &0.53&10.26&12.28&126.99&0.67&0.59&56.52&43.21&32.19&0.30&0.50\\ 

LDM-4-G (cfg=1.5) & 178k$\times$1200 & 
3.60 & 5.12 & 247.67 & \textbf{0.87} & 0.48 &     
10.04 & 11.47 & 119.56 & 0.65 & 0.61 &    
29.67 & 26.33 & 57.71 & 0.44 & \textbf{0.61} \\

MDT-G$^{\dag}$ (cfg=3.8,s=4) & 6500k$\times$256 & 
\textbf{1.79} & \textbf{4.57} & \textbf{283.01} & 0.81 & \textbf{0.61} &     
135.6 & 73.08 & 9.35 & 0.15 & 0.20 &    
124.9 & 70.69 & 13.38 & 0.13 & 0.42 \\

DiT-XL/2-G (cfg=1.50) & 7000k$\times$256 & 
2.27 & 4.60 & 278.24 & 0.83 & 0.57 &     
20.14 & 30.50 & 97.28 & 0.49 & \textbf{0.67} &   
107.2 & 68.89 & 15.48 & 0.12 & 0.52 \\

% DiT-XL/2-G (cfg=1.50)  & 1800k$\times$256  &
% 2.66 & 4.62 & 238.90 & 0.83 & 0.56 & 
% 25.01 & 47.72 & 80.19 & 0.47 & 0.62 &
% 107.2 & 94.02 & 13.25 & 0.14 & 0.45 \\

% FiT-XL/2-G (cfg=1.50)  & 1800k$\times$256 
% &4.27&9.99&249.72&0.84&0.51
% &5.81&9.93&186.54&0.73&0.55
% &17.68&21.02&105.17&0.57&0.52\\

\baseline{FiT-XL/2-G$^{\ast}$ (cfg=1.50)}  & \baseline{1800k$\times$256} 
&\baseline{4.27}&\baseline{9.99}&\baseline{249.72}&\baseline{0.84}&\baseline{0.51}
&\baseline{\textbf{5.74}}&\baseline{\textbf{10.05}}&\baseline{\textbf{190.14}}&\baseline{\textbf{0.74}}&\baseline{0.55}
&\baseline{\textbf{16.81}}&\baseline{\textbf{20.62}}&\baseline{\textbf{110.93}}&\baseline{\textbf{0.57}}&\baseline{0.52}\\

\bottomrule[1.2pt]

\end{tabular}
\end{adjustbox}

\caption{Benchmarking class-conditional image generation with in-distribution resolution on \textit{ImageNet} dataset. ``-G'' denotes
the results with classifier-free guidance. $^{\dag}$: MDT-G adpots an improved classifier-free guidance strategy~\cite{gao2023masked}: $w_t=(1-\cos\pi (\frac{t}{t_{max}})^s)w/2$, where $w=3.8$ is the maximum guidance scale and $s=4$ is the controlling factor.
$^{\ast}$: FiT-XL/2-G adopts VisionNTK for resolution extrapolation.
}
\vskip -0.1in
\label{tab:in1k_id}
\end{table*}
\begin{table*}[t]
\centering
% \addtolength{\tabcolsep}{-1pt}

\begin{adjustbox}{max width=1.\textwidth}
\begin{tabular}{l|c|ccccc|ccccc|ccccc}
\toprule[1.2pt]
\multirow{2}*{Method}&\multirow{2}*{Train Cost}&\multicolumn{5}{c|}{320$\times$320 (1:1)}&\multicolumn{5}{c|}{224$\times$448 (1:2)}&\multicolumn{5}{c}{160$\times$480 (1:3)} \\

& 

& \textbf{FID$\downarrow$} & \textbf{sFID$\downarrow$} & \textbf{IS$\uparrow$} & \textbf{Prec.$\uparrow$} & \textbf{Rec.$\uparrow$}
& \textbf{FID$\downarrow$} & \textbf{sFID$\downarrow$} & \textbf{IS$\uparrow$} & \textbf{Prec.$\uparrow$} & \textbf{Rec.$\uparrow$}  
& \textbf{FID$\downarrow$} & \textbf{sFID$\downarrow$} & \textbf{IS$\uparrow$} & \textbf{Prec.$\uparrow$} & \textbf{Rec.$\uparrow$}  \\
% & \textbf{FID} & \textbf{sFID} & \textbf{IS} & \textbf{Prec.} & \textbf{Rec.}
% & \textbf{FID} & \textbf{sFID} & \textbf{IS} & \textbf{Prec.} & \textbf{Rec.}  
% & \textbf{FID} & \textbf{sFID} & \textbf{IS} & \textbf{Prec.} & \textbf{Rec.}  \\

\midrule

U-ViT-H/2-G (cfg=1.4) & 500k$\times$1024  &7.65&16.30&208.01&0.72&\textbf{0.54}&67.10&42.92&45.54&0.30 &\textbf{0.49}&95.56&44.45&24.01&0.19&0.47  \\

ADM-G,U & 1980k$\times$256 & 9.39 &\textbf{9.01}&161.95&0.74&0.50&11.34&\textbf{14.50}&146.00&0.71&\textbf{0.49}&23.92&25.55&80.73&0.57& \textbf{0.51} \\ 

LDM-4-G (cfg=1.5) & 178k$\times$1200 & 
6.24 & 13.21 & 220.03 & \textbf{0.83} & 0.44 &     
8.55 & 17.62 & 186.25 & \textbf{0.78} & 0.44 &    
19.24 & \textbf{20.25} & 99.34 & 0.59 & 0.50 \\

MDT-G$^{\dag}$ (cfg=3.8,s=4) & 6500k$\times$256 & 
383.5 & 136.5 & 4.24 & 0.01 & 0.04 &     
365.9 & 142.8 & 4.91 & 0.01 & 0.05 &    
276.7 & 138.1 & 7.20 & 0.03 & 0.09 \\

DiT-XL/2-G (cfg=1.50) & 7000k$\times$256 & 
9.98 & 23.57 & 225.72 & 0.73 & 0.48 &     
94.94 & 56.06 & 35.75 & 0.23 & 0.46 &    
140.2 & 79.60 & 14.70 & 0.094 & 0.45 \\

% DiT-XL/2-G (cfg=1.50) & 1800k$\times$256 &
% 9.79 & 23.38 & 214.64 & 0.75 & 0.44 &
% 77.77 & 56.14 & 43.41 & 0.28 & 0.39 &
% 117.1 & 88.87 & 16.52 & 0.12 & 0.40 \\

% FiT-XL/2-G (cfg=1.50) & 1800k$\times$256 
% &5.14&14.08&252.41&0.82&0.47&7.73&19.05&213.34&0.75&0.47&15.87 & 22.14 &128.32 &0.62 &0.47
% \\

\baseline{FiT-XL/2-G$^{\ast}$ (cfg=1.50)}  & \baseline{1800k$\times$256}
&\baseline{\textbf{5.42}}&\baseline{15.41}&\baseline{\textbf{252.65}}&\baseline{0.81}&\baseline{0.47}&\baseline{\textbf{7.90}}&\baseline{19.63}&\baseline{\textbf{215.29}}&\baseline{0.75}&\baseline{0.47}&\baseline{\textbf{15.72}}&\baseline{22.57}&\baseline{\textbf{132.76}}&\baseline{\textbf{0.62}}&\baseline{0.47}
\\

\bottomrule[1.2pt]

\end{tabular}
\end{adjustbox}

\caption{Benchmarking class-conditional image generation with out-of-distribution resolution on \textit{ImageNet} dataset. ``-G'' denotes the results with classifier-free guidance. $^{\dag}$: MDT-G adopts an aforementioned improved classifier-free guidance strategy.
$^{\ast}$: FiT-XL/2-G adopts VisionNTK for resolution extrapolation. Our FiT model achieves state-of-the-art performance across all the resolutions and aspect ratios, demonstrating a strong extrapolation capability.}

\vskip -0.15in

\label{tab:in1k_ood}
\end{table*}
\begin{table*}[t]
\centering
% \addtolength{\tabcolsep}{-1pt}
% \setlength\tabcolsep{1pt}
\begin{adjustbox}{max width=1.0\textwidth}
\begin{tabular}{l|ccc|ccccc|ccccc|ccccc}
\toprule[1.2pt]
\multirow{2}*{Arch.} &\multirow{2}*{Pos. Embed.} &\multirow{2}*{FFN} &\multirow{2}*{Train}&\multicolumn{5}{c|}{256$\times$256 (i.d.)} &\multicolumn{5}{c|}{160$\times$320 (i.d.)} &\multicolumn{5}{c}{224$\times$448 (o.o.d.)}\\
 & & & 
 
& \textbf{FID$\downarrow$} & \textbf{sFID$\downarrow$} & \textbf{IS$\uparrow$} & \textbf{Prec.$\uparrow$} & \textbf{Rec.$\uparrow$}
& \textbf{FID$\downarrow$} & \textbf{sFID$\downarrow$} & \textbf{IS$\uparrow$} & \textbf{Prec.$\uparrow$} & \textbf{Rec.$\uparrow$}  
& \textbf{FID$\downarrow$} & \textbf{sFID$\downarrow$} & \textbf{IS$\uparrow$} & \textbf{Prec.$\uparrow$} & \textbf{Rec.$\uparrow$}  \\
% & \textbf{FID} & \textbf{sFID} & \textbf{IS} & \textbf{Prec.} & \textbf{Rec.}
% & \textbf{FID} & \textbf{sFID} & \textbf{IS} & \textbf{Prec.} & \textbf{Rec.}  
% & \textbf{FID} & \textbf{sFID} & \textbf{IS} & \textbf{Prec.} & \textbf{Rec.}  \\

\midrule

DiT-B & Abs. PE & MLP & Fixed & 
44.83 & \textbf{8.49} & 32.05 & 0.48 & \textbf{0.63} &
91.32 & 66.66 & 14.02 & 0.21 & 0.45 &
109.1 & 110.71 & 14.00 & 0.18 & 0.31 \\

\midrule
\color{codegray}{Config A} & Abs. PE & MLP & Flexible  & 43.34 & 11.11 & 32.23 & 0.48 & 0.61 & 50.51 & 10.36 & 25.26 & 0.42 & 0.60 & 52.55 & 16.05 & 28.69 & 0.42 & \textbf{0.58} \\
 
\color{codegray}{Config B} & Abs. PE & SwiGLU & Flexible  & 41.75 & 11.53 & 34.55 & 0.49 & 0.61 & 48.66 & 10.65 & 26.76 & 0.41 & 0.60 & 52.34 & 17.73 & 30.01 & 0.41 & 0.57 \\
 
\color{codegray}{Config C} & Abs. PE + 2D RoPE & MLP & Flexible  & 39.11 & 10.79 & 36.35 & 0.51 & 0.61 & 46.71 & 10.32 & 27.65 & \textbf{0.44} & 0.61 & 46.60 & \textbf{15.84} & 33.99 & 0.46 & \textbf{0.58} \\
 
\color{codegray}{Config D} & 2D RoPE & MLP & Flexible  & 37.29 & 10.62 & 38.34 & \textbf{0.53} & 0.61 & 45.06 & \textbf{9.82} & 28.87 & 0.43 & 0.62 & 46.16 & 23.72 & 35.28 & 0.46 & 0.55 \\
% \color{Maroon}{FiT-B}
FiT-B & 2D RoPE & SwiGLU & Flexible  & \textbf{36.36} & 11.08 & \textbf{40.69} & 0.52 & 0.62 & \textbf{43.96} & 10.26 & \textbf{30.45} & 0.43 & \textbf{0.62} & \textbf{44.67} & 24.09 & \textbf{37.10} & \textbf{0.49} & 0.53 \\
\bottomrule[1.2pt]

\end{tabular}
\end{adjustbox}

\caption{Ablation results from DiT-B/2 to FiT-B/2 at  400K training steps without using classifier-free guidance. }
\vskip -0.1in
\label{tab:ablation}
\end{table*}
\begin{table*}[t]
\centering
% \addtolength{\tabcolsep}{-1pt}

\begin{adjustbox}{max width=1.\textwidth}
\begin{tabular}{l|ccccc|ccccc|ccccc}
\toprule[1.2pt]
\multirow{2}*{Method}&\multicolumn{5}{c|}{320$\times$320 (1:1)}&\multicolumn{5}{c|}{224$\times$448 (1:2)}&\multicolumn{5}{c}{160$\times$480 (1:3)} \\

& \textbf{FID$\downarrow$} & \textbf{sFID$\downarrow$} & \textbf{IS$\uparrow$} & \textbf{Prec.$\uparrow$} & \textbf{Rec.$\uparrow$}
& \textbf{FID$\downarrow$} & \textbf{sFID$\downarrow$} & \textbf{IS$\uparrow$} & \textbf{Prec.$\uparrow$} & \textbf{Rec.$\uparrow$}  
& \textbf{FID$\downarrow$} & \textbf{sFID$\downarrow$} & \textbf{IS$\uparrow$} & \textbf{Prec.$\uparrow$} & \textbf{Rec.$\uparrow$}  \\
% & \textbf{FID} & \textbf{sFID} & \textbf{IS} & \textbf{Prec.} & \textbf{Rec.}
% & \textbf{FID} & \textbf{sFID} & \textbf{IS} & \textbf{Prec.} & \textbf{Rec.}  
% & \textbf{FID} & \textbf{sFID} & \textbf{IS} & \textbf{Prec.} & \textbf{Rec.}  \\

\midrule

DiT-B & 95.47 & 108.68 & 18.38 & 0.26 & 0.40 &     
109.1 & 110.71 & 14.00 & 0.18 & 0.31 &    
143.8 & 122.81 & 8.93 & 0.073 & 0.20 \\

DiT-B + EI & 81.48 & 62.25 & 20.97 & 0.25 & 0.47 &    
133.2 & 72.53 & 11.11 & 0.11 & 0.29 &    
160.4 & 93.91 & 7.30 & 0.054 & 0.16 \\

DiT-B + PI & 72.47 & 54.02 & 24.15 & 0.29 & 0.49 &    
133.4 & 70.29 & 11.73 & 0.11 & 0.29 &    
156.5 & 93.80 & 7.80 & 0.058 & 0.17 \\
\midrule
FiT-B & 61.35 & \textbf{30.71} & 31.01 & 0.41 & 0.53 & 44.67 & \textbf{24.09} & 37.1 & 0.49 & 0.52 & 56.81 & \textbf{22.07} & 25.25 & \textbf{0.38} & 0.49 \\

FiT-B + PI & 65.76 & 65.45 & 29.32 & 0.32 & 0.45 & 175.42 & 114.39 & 8.45 & 0.14 & 0.06 & 224.83 & 123.45 & 5.89 & 0.02 & 0.06 \\

% FiT-B + PartNTK & 49.40 & 38.21 & 40.95 & 0.47 & 0.53 & 103.18 & 79.13 & 24.32 & 0.23 & 0.38 & 122.74 & 71.19 & 17.88 & 0.17 & 0.32 \\

FiT-B + YaRN & 44.76 & 38.04 & 44.70 & 0.51 & 0.51 & 82.19 & 75.48 & 29.68  & 0.40 & 0.29 & 104.06 & 72.97 & 20.76 & 0.21 & 0.31 \\

FiT-B + NTK & 57.31 & 31.31 & 33.97 & 0.43 & 0.55 & 45.24 & 29.38 & 38.84 & 0.47 & 0.52 & 59.19 & 26.54 & 26.01 & 0.36 & 0.49\\

\midrule

FiT-B + \textbf{VisionYaRN} & \textbf{44.76} & 38.04 & \textbf{44.70} & \textbf{0.51} & 0.51 & \textbf{41.92} & 42.79 & \textbf{45.87} & \textbf{0.50} & 0.48 & 62.84 & 44.82 & \textbf{27.84} & 0.36 & 0.42 \\

FiT-B + \textbf{VisionNTK} & 57.31 & 31.31 & 33.97 & 0.43 & \textbf{0.55} & 43.84 & 26.25 & 39.22 & 0.48 & \textbf{0.52} & \textbf{56.76} & 24.18 & 26.40 & 0.37 & \textbf{0.49} \\
\bottomrule[1.2pt]

\end{tabular}
\end{adjustbox}

\caption{Benchmarking class-conditional image generation with out-of-distribution resolution on ImageNet. The FiT-B/2 and DiT-B/2 at 400K training steps are adopted in this experiment. Metrics are calculated without using classifier-free guidance. YaRN and NTK mean the vanilla implementation of such two methods. Our FiT-B/2 demonstrates stable extrapolation performance, which can be further improved combined with VisionNTK and VisionYaRN methods.}
\vskip -0.15in
\label{tab:ablation_in1k_ood}
\end{table*}
% \input{table/ablation_extrapolation_hard}
% \input{table/transfer_high_resolution}

% \xihui{Didn't mention the dataset? ImageNet?}

\subsection{FiT Implementation}

We present the implementation details of FiT, including model architecture, training details, and evaluation metrics.

\noindent\textbf{Model architecture.}
We follow DiT-B and DiT-XL to set the same layers, hidden size, and attention heads for base model FiT-B and xlarge model FiT-XL. 
As DiT reveals stronger synthesis performance when using a smaller patch size, we use a patch size p=2, denoted by FiT-B/2 and FiT-XL/2. 
FiT adopts the same off-the-shelf pre-trained VAE~\cite{esser2021taming} as DiT provided by the Stable Diffusion~\cite{rombach2022high} to encode/decode the image/latent tokens. 
The VAE encoder has a downsampling ratio of $1/8$ and a feature channel dimension of $4$.
An image of size $160\times320\times3$ is encoded into latent codes of size $20\times40\times4$.
The latent codes of size $20\times40\times4$ are patchified into latent tokens of length $L=10\times20=200$.

\noindent\textbf{Training details.} We train class-conditional latent FiT models under predetermined maximum resolution limitation, $HW\leqslant256^2$ (equivalent to token length $L\leq256$), on the \textit{ImageNet}~\cite{deng2009imagenet} dataset.
We down-resize the high-resolution images to meet the $HW\leqslant256^2$ limitation while maintaining the aspect ratio.
We follow DiT to use Horizontal Flip Augmentation.
We use the same training setting as DiT: a constant learning rate of $1\times10^{-4}$ using AdamW~\cite{loshchilov2017decoupled}, no weight decay, and a batch size of $256$.
Following common practice in the generative modeling literature, we adopt an exponential moving average (EMA) of model weights over training with a decay of 0.9999. 
All results are reported using the EMA model. 
We retain the same diffusion hyper-parameters as DiT.
 
\noindent\textbf{Evaluation details and metrics.} We evaluate models with some commonly used metrics, \textit{i.e.} Fre’chet Inception Distance (FID)~\cite{heusel2017fid}, sFID~\cite{nash2021sfid}, Inception Score (IS)~\cite{salimans2017inceptionscore}, improved Precision and Recall~\cite{kynk2017precisionrecall}. 
For fair comparisons, we follow DiT to use the TensorFlow evaluation from ADM~\cite{dhariwal2021diffusion} and report FID-50K with 250 DDPM sampling steps. 
FID is used as the major metric as it measures both diversity and fidelity.
We additionally report IS, sFID, Precision, and Recall as secondary metrics.
For FiT architecture experiment (\cref{subsec:exp_fit_architecture}) and resolution extrapolation ablation experiment (\cref{subsec:exp_extrapolation_design}), we report the results without using classifier-free guidance~\cite{ho2021cfg}.

\noindent\textbf{Evaluation resolution.} Unlike previous work that mainly conducted experiments on a fixed aspect ratio of $1:1$, we conducted experiments on different aspect ratios, which are $1:1$, $1:2$, and $1:3$, respectively.
On the other hand, we divide the experiment into resolution within the training distribution and resolution out of the training distribution.
For the resolution in distribution, we mainly use $256 \times 256$ (1:1), $160 \times 320$ (1:2), and $128 \times 384$ (1:3) for evaluation, with $256$, $200$, $192$ latent tokens respectively. 
All token lengths are smaller than or equal to 256, leading to respective resolutions within the training distribution.
For the resolution out of distribution, we mainly use $320 \times 320$ (1:1), $224 \times 448$ (1:2), and $160 \times 480$ (1:3) for evaluation, with $400$, $392$, $300$ latent tokens respectively.
All token lengths are larger than 256, resulting in the resolutions out of training distribution.
Through such division, we holistically evaluate the image synthesis and resolution extrapolation ability of FiT at various resolutions and aspect ratios.

\subsection{FiT Architecture Design}
\label{subsec:exp_fit_architecture}
In this part, we conduct an ablation study to verify the architecture designs in FiT. We report the results of various variant FiT-B/2 models at 400K training steps and use FID-50k, sFID, IS, Precision, and Recall as the evaluation metrics.
We conduct experiments at three different resolutions: $256\times256$, $160\times320$, and $224\times448$. 
These resolutions are chosen to encompass different aspect ratios, as well as to include resolutions both in and out of the distribution.

\noindent\textbf{Flexible training vs. Fixed training.} \textit{Flexible training pipeline significantly improves the performance across various resolutions.} 
This improvement is evident not only within the in-distribution resolutions but also extends to resolutions out of the training distribution, as shown in Tab.~\ref{tab:ablation}.
\textcolor{codegray}{Config A} is the original DiT-B/2 model only with flexible training, which slightly improves the performance (\textcolor{coolblack}{-1.49} FID) compared with DiT-B/2 with fixed resolution training at $256\times256$ resolution.
\textcolor{codegray}{Config A} demonstrates a significant performance improvement through flexible training. Compared to DiT-B/2, FID scores are reduced by \textcolor{coolblack}{40.81} and \textcolor{coolblack}{56.55} at resolutions $160\times320$ and $224\times448$, respectively.
% Specifically, at these resolutions, the performance enhancements of Config A were 40.81 and 56.55, respectively.(\textcolor{codegray}{-40.81 FID}, \textcolor{codegray}{-56.55 FID}) 
% So FiT uses flexible training instead of fixed-resolution training.

\noindent\textbf{SwiGLU vs. MLP.} \textit{SwiGLU slightly improves the performance across various resolutions, compared to MLP.}
\textcolor{codegray}{Config B} is the FiT-B/2 flexible training model replacing MLP with SwiGLU.
Compared to \textcolor{codegray}{Config A}, \textcolor{codegray}{Config B} demonstrates notable improvements across various resolutions. 
Specifically, for resolutions of $256\times256$, $160\times320$, and $224\times448$, \textcolor{codegray}{Config B} reduces the FID scores by \textcolor{coolblack}{1.59}, \textcolor{coolblack}{1.85}, and \textcolor{coolblack}{0.21} in Tab.~\ref{tab:ablation}, respectively.
So FiT uses SwiGLU in FFN.

\noindent\textbf{2D RoPE vs. Absolute PE.} \textit{2D RoPE demonstrates greater efficiency compared to absolute position encoding, and it possesses significant extrapolation capability across various resolutions.}
\textcolor{codegray}{Config D} is the FiT-B/2 flexible training model replacing absolute PE with 2D RoPE.
For resolutions within the training distribution, specifically $256\times256$ and $160\times320$, \textcolor{codegray}{Config D} reduces the FID scores by \textcolor{coolblack}{6.05}, and \textcolor{coolblack}{5.45} in Tab.~\ref{tab:ablation}, compared to \textcolor{codegray}{Config A}.
For resolution beyond the training distribution, $224\times448$, \textcolor{codegray}{Config D} shows significant extrapolation capability (\textcolor{coolblack}{-6.39} FID) compared to \textcolor{codegray}{Config A}.
\textcolor{codegray}{Config C} retains both absolute PE and 2D RoPE.
However, in a comparison between \textcolor{codegray}{Config C} and \textcolor{codegray}{Config D}, we observe that \textcolor{codegray}{Config C} performs worse.
For resolutions of 256x256, 160x320, and 224x448, \textcolor{codegray}{Config C} increases FID scores of \textcolor{coolblack}{1.82}, \textcolor{coolblack}{1.65}, and \textcolor{coolblack}{0.44}, respectively, compared to \textcolor{codegray}{Config D}.
Therefore, only 2D RoPE is used for positional embedding in our implementation.

\noindent\textbf{Putting it together.} \textit{FiT demonstrates significant and comprehensive superiority across various resolution settings, compared to original DiT.}
FiT has achieved state-of-the-art performance across various configurations. 
Compared to DiT-B/2, FiT-B/2 reduces the FID score by \textcolor{coolblack}{8.47} on the most common resolution of $256\times256$ in Tab.~\ref{tab:ablation}. 
Furthermore, FiT-B/2 has made significant performance gains at resolutions of $160\times320$ and $224\times448$, decreasing the FID scores by \textcolor{coolblack}{47.36} and \textcolor{coolblack}{64.43}, respectively.

\subsection{FiT Resolution Extrapolation Design}
\label{subsec:exp_extrapolation_design}

In this part, we adopt the DiT-B/2 and FiT-B/2 models at 400K training steps to evaluate the extrapolation performance on three out-of-distribution resolutions: $320\times320$, $224\times448$ and $160\times480$. Direct extrapolation does not perform well on larger resolution out of training distribution. So we conduct a comprehensive benchmarking analysis focused on positional embedding interpolation methods.

\noindent \textbf{PI and EI.} PI (Position Interpolation) and EI (Embedding Interpolation) are two baseline positional embedding interpolation methods for resolution extrapolation. PI linearly down-scales the inference position coordinates to match the original coordinates. EI resizes the positional embedding with bilinear interpolation\footnote{torch.nn.functional.interpolate(pe, (h,w), method='bilinear')}. Following ViT~\cite{dosovitskiy2020image}, EI is used for absolute positional embedding.

\noindent \textbf{NTK and YaRN.} We set the scale factor to $s=\max({\max(H_{\text{test}}, W_{\text{test}})}/{\sqrt{256}})$ and adopt the vanilla implementation of the two methods, as in \cref{subsec:method_preliminary}. For YaRN, we set $\alpha=1, \beta=32$ in \cref{eq:1d_partntk_ramp}.

\noindent \textbf{VisionNTK and VisionYaRN.} These two methods are defined detailedly in \cref{def:vision_ntk,def:vision_yarn}. Note that when the aspect ratio equals one, the VisionNTK and VisionYaRN are equivalent to NTK and YaRN, respectively.

\noindent \textbf{Analysis.} We present in Tab.~\ref{tab:ablation_in1k_ood} that our FiT-B/2 shows stable performance when directly extrapolating to larger resolutions. When combined with PI, the extrapolation performance of FiT-B/2 at all three resolutions decreases. When combined with YaRN, the FID score reduces by \textcolor{coolblack}{16.77} on $320\times320$, but the performance on $224\times448$ and $168\times480$ descends. Our VisionYaRN solves this dilemma and reduces the FID score by \textcolor{coolblack}{40.27} on $224\times448$ and by \textcolor{coolblack}{41.22} at $160\times480$ compared with YaRN. NTK interpolation method demonstrates stable extrapolation performance but increases the FID score slightly at $224\times448$ and $160\times480$ resolutions. Our VisionNTK method alleviates this problem and exceeds the performance of direct extrapolation at all three resolutions. In conclusion, our FiT-B/2 has a strong extrapolation ability, which can be further enhanced when combined with VisionYaRN and VisionNTK methods.

However, DiT-B/2 demonstrates poor extrapolation ability. When combined with PI, the FID score achieves \textcolor{coolblack}{72.47} at $320\times320$ resolution, which still falls behind our FiT-B/2. At $224\times448$ and $160\times480$ resolutions, PI and EI interpolation methods cannot improve the extrapolation performance.

\subsection{FiT In-Distribution Resolution Results}
\label{subsec:exp_in_distribution}

Following our former analysis, we train our highest Gflops model, FiT-XL/2, for 1.8M steps. 
% In this part, we conduct experiments % ablation to verify the architecture designs in FiT. 
% We use FID-50k, sFID, IS, Precision, and Recall as the evaluation metric. 
We conduct experiments to evaluate the performance of FiT at three different in distribution resolutions: $256\times256$, $160\times320$, and $128\times384$. 
% As shown in Tab.~\ref{tab:in1k_id},
We show samples from the FiT in Fig~\ref{fig:teaser}, and we compare against some state-of-the-art class-conditional generative models: BigGAN~\cite{brock2018large}, StyleGAN-XL~\cite{sauer2022stylegan}, MaskGIT~\cite{chang2022maskgit}, CDM~\cite{ho2022cascaded}, U-ViT~\cite{bao2023all}, ADM~\cite{dhariwal2021diffusion}, LDM~\cite{rombach2022high}, MDT~\cite{gao2023masked}, and DiT~\cite{peebles2023scalable}. 
When generating images of $160\times320$ and $128\times384$ resolution, we adopt PI on the positional embedding of the DiT model, as stated in \cref{subsec:exp_extrapolation_design}. EI is employed in the positional embedding of U-ViT and MDT models, as they use learnable positional embedding. ADM and LDM can directly synthesize images with resolutions different from the training resolution. 

As shown in Tab.~\ref{tab:in1k_id}, FiT-XL/2 outperforms all prior diffusion models, decreasing the previous best FID-50K of \textcolor{coolblack}{6.93} achieved by U-ViT-H/2-G to \textcolor{coolblack}{5.74} at $160\times320$ resolution.
For $128\times384$ resolution, FiT-XL/2 shows significant superiority, decreasing the previous SOTA FID-50K of \textcolor{coolblack}{29.67} to \textcolor{coolblack}{16.81}.
The FID score of FiT-XL/2 increases slightly at $256\times256$ resolution, compared to other models that have undergone longer training steps.
% These resolutions are chosen to encompass different aspect ratios, as well as to include resolutions both in and out of the distribution.

\subsection{FiT Out-Of-Distribution Resolution Results}
\label{subsec:exp_out_of_distribution}

We evaluate our FiT-XL/2 on three different out-of-distribution resolutions: $320\times320$, $224\times448$, and $160\times480$ and compare against some SOTA class-conditional generative models: U-ViT, ADM, LDM-4, MDT, and DiT. PI is employed in DiT, while EI is adopted in U-ViT and MDT, as in \cref{subsec:exp_in_distribution}. U-Net-based methods, such as ADM and LDM-4 can directly generate images with resolution out of distribution. As shown in \cref{tab:in1k_ood}, FiT-XL/2 achieves the best FID-50K and IS, on all three resolutions, indicating its outstanding extrapolation ability. In terms of other metrics, as sFID, FiT-XL/2 demonstrates competitive performance. 

LDMs with transformer backbones are known to have difficulty in generating images out of training resolution, such as DiT, U-ViT, and MDT. More seriously, MDT has almost no ability to generate images beyond the training resolution. We speculate this is because both learnable absolute PE and learnable relative PE are used in MDT. DiT and U-ViT show a certain degree of extrapolation ability and achieve FID scores of \textcolor{coolblack}{9.98} and \textcolor{coolblack}{7.65} respectively at 320x320 resolution. However, when the aspect ratio is not equal to one, their generation performance drops significantly, as $224\times448$ and $160\times480$ resolutions. Benefiting from the advantage of the local receptive field of the Convolution Neural Network, ADM and LDM show stable performance at these out-of-distribution resolutions. Our FiT-XL/2 solves the problem of insufficient extrapolation capabilities of the transformer in image synthesis. At $320\times320$, $224\times448$, and $160\times480$ resolutions, FiT-XL/2 exceeds the previous SOTA LDM on FID-50K by \textcolor{coolblack}{0.82}, \textcolor{coolblack}{0.65}, and \textcolor{coolblack}{3.52} respectively.
\section{Conclusion}

In this work,  we aim to contribute to the ongoing research on flexible generating arbitrary resolutions and aspect ratio.
We propose Flexible Vision Transformer (FiT) for diffusion model, a refined transformer architecture with flexible training pipeline specifically designed for generating images with arbitrary resolutions and aspect ratios.
FiT surpasses all previous models, whether transformer-based or CNN-based, across various resolutions. 
With our resolution extrapolation method, VisionNTK, the performance of FiT has been significantly enhanced further.

\nocite{langley00}

\bibliography{main}
\bibliographystyle{icml2024}

%%%%%%%%%%%%%%%%%%%%%%%%%%%%%%%%%%%%%%%%%%%%%%%%%%%%%%%%%%%%%%%%%%%%%%%%%%%%%%%
%%%%%%%%%%%%%%%%%%%%%%%%%%%%%%%%%%%%%%%%%%%%%%%%%%%%%%%%%%%%%%%%%%%%%%%%%%%%%%%
% APPENDIX
%%%%%%%%%%%%%%%%%%%%%%%%%%%%%%%%%%%%%%%%%%%%%%%%%%%%%%%%%%%%%%%%%%%%%%%%%%%%%%%
%%%%%%%%%%%%%%%%%%%%%%%%%%%%%%%%%%%%%%%%%%%%%%%%%%%%%%%%%%%%%%%%%%%%%%%%%%%%%%%
\newpage
\appendix
\onecolumn
\section{Experimentin Setups}

We provide detailed network configurations and performance of all models, which are listed in Tab.~\ref{tab:config}.
% and training costs for MDT under different model scales
% are listed in Tab. 6. In comparison to DiT baselines, MDT
% introduces a negligible extra inference parameters and costs

\begin{table}[ht]
\centering
\begin{adjustbox}{max width=1.\textwidth}
\begin{tabular}{lccccccccc}
\toprule[1.2pt]
Models & Layers & Dim. & Head Num. & Patch Size & Max Token Length & Training Steps & Batch Size & Learning Rate & FID-50K \\ \midrule[1.2pt]
% \multicolumn{6}{c}{Network configurations of FiT models.} \\
DiT-B/2& 12 & 768 & 12 & 2 & 256 & 400K & 256 & $1\times10^{-4}$&44.83 \\ 
\textcolor{codegray}{Config A}& 12 & 768 & 12 & 2 & 256 & 400K & 256 & $1\times10^{-4}$&43.34 \\
\textcolor{codegray}{Config B}& 12 & 768 & 12 & 2 & 256 & 400K & 256 & $1\times10^{-4}$&41.75 \\
\textcolor{codegray}{Config C}& 12 & 768 & 12 & 2 & 256 & 400K & 256 & $1\times10^{-4}$&39.11 \\
\textcolor{codegray}{Config D}& 12 & 768 & 12 & 2 & 256 & 400K & 256 & $1\times10^{-4}$&37.29 \\
FiT-B/2& 12 & 768 & 12 & 2 & 256 & 400K & 256 & $1\times10^{-4}$&36.36 \\ \midrule
FiT-XL/2-G& 28 & 1152 & 16 & 2 & 256 & 1800K & 256 & $1\times10^{-4}$ &4.27 \\ \bottomrule[1.2pt]
\end{tabular}
\end{adjustbox}
\caption{Network configurations and performance of all models.}
\label{tab:config}
\end{table}

We use the same ft-EMA VAE\footnote{The model is downloaded in \url{https://huggingface.co/stabilityai/sd-vae-ft-ema}} with DiT, which is provided by the Stable Diffusion to encode/decode the image/latent tokens by default. The metrics are calculated using the ADM TensorFlow evaluation Suite\footnote{\url{https://github.com/openai/guided-diffusion/tree/main/evaluations}}.

\section{Detailed Attention Score with 2D RoPE and decoupled 2D-RoPE.}
\label{detail_attn_score}

2D RoPE defines a vector-valued complex function $f(\mathbf{x}, h_m, w_m)$ in ~\cref{eq:2d_rope} as follows:
\begin{equation}
\begin{aligned}
f(\mathbf{x}, h_m, w_m) &=  \left[ 
(x_0 + ix_1)e^{ih_m\theta_0}, (x_2 + ix_3)e^{ih_m\theta_1}, \ldots, (x_{d/2-2} + ix_{d/2-1})e^{ih_m\theta_{d/4-1}}, 
\right. \\
& \quad  \left.
(x_{d/2} + ix_{d/2+1})e^{iw_m\theta_0}, (x_{d/2+2} + ix_{d/2+3})e^{iw_m\theta_1}, \ldots, (x_{d-2} + ix_{d-1})e^{iw_m\theta_{d/4-1}} 
\right]^T .
\end{aligned} 
\end{equation}

The self-attention score $A_n$ injected with 2D RoPE in \cref{eq:2d_rope_attn} is detailed defined as follows:
\begin{equation}
\begin{aligned}
% Q_i K_i^T 
% a((x_0,y_0), (x_1,y_1)) &= \mathrm{Re} \langle f(q,(x_0,y_0)), f(k,(x_1,y_1)) \rangle \\
A_n =& \mathrm{Re}\langle f_q(\mathbf{q}_m, h_m, w_m), f_k(\mathbf{k}_n, h_n, w_n)\rangle \\
=& \mathrm{Re} 
\left[ 
\sum^{d/4-1}_{j=0} (q_{2j}+iq_{2j+1})(k_{2j}-ik_{2j+1})e^{i(h_m-h_n)\theta_j} + \right. 
\left. \sum^{d/4-1}_{j=0} (q_{2j}+iq_{2j+1})(k_{2j}-ik_{2j+1})e^{i(w_m-w_n)\theta_j} 
\right] \\
=&\sum^{d/4-1}_{j=0} [ (q_{2j}k_{2j} + q_{2j+1}k_{2j+1})\cos((h_m-h_n)\theta_j) +  
(q_{2j}k_{2j+1} - q_{2j+1}k_{2j})\sin((h_m-h_n)\theta_j)]+ \\
&\sum^{d/4-1}_{j=0} [ (q_{2j}k_{2j} + q_{2j+1}k_{2j+1})\cos((w_m-w_n)\theta_j) + 
(q_{2j}k_{2j+1} - q_{2j+1}k_{2j})\sin((w_m-w_n)\theta_j)],\\
\end{aligned} 
\end{equation}
where 2-D coordinates of width and height as $\{(w, h) \Big| 1\leqslant w \leqslant W, 1 \leqslant h \leqslant H\}$, the subscripts of $q$ and $k$ denote the dimensions of the attention head, $\theta^n=10000^{-2n/d}$.
There is no cross-term between $h$ and $w$ in 2D-RoPE and attention score $A_n$, so we can further decouple the rotary frequency as $\Theta_h=\{\theta^h_d=b_h^{-2d/|D|}, 1\leqslant d\leqslant \frac{|D|}{2}\}$ and $\Theta_w=\{\theta^w_d=b_w^{-2d/|D|}, 1\leqslant d\leqslant \frac{|D|}{2}\}$, resulting in the decoupled 2D-RoPE, as follows:

\begin{equation}
\begin{aligned}
A_n=&\sum^{d/4-1}_{j=0} [ (q_{2j}k_{2j} + q_{2j+1}k_{2j+1})\cos((h_m-h_n)\theta^h_j) + (q_{2j}k_{2j+1} - q_{2j+1}k_{2j})\sin((h_m-h_n)\theta^h_j)]+ \\
&\sum^{d/4-1}_{j=0} [ (q_{2j}k_{2j} + q_{2j+1}k_{2j+1})\cos((w_m-w_n)\theta^w_j) + (q_{2j}k_{2j+1} - q_{2j+1}k_{2j})\sin((w_m-w_n)\theta^w_j)] \\
=&\mathrm{Re} 
\left[ 
\sum^{d/4-1}_{j=0} (q_{2j}+iq_{2j+1})(k_{2j}-ik_{2j+1})e^{i(h_m-h_n)\theta_j^h} + \sum^{d/4-1}_{j=0} (q_{2j}+iq_{2j+1})(k_{2j}-ik_{2j+1})e^{i(w_m-w_n)\theta_j^w} 
\right] \\
=&\mathrm{Re}\langle \hat{f}_q(\mathbf{q}_m, h_m, w_m), \hat{f}_k(\mathbf{k}_n, h_n, w_n)\rangle.\\
\end{aligned} 
\end{equation}

So we can reformulate the vector-valued complex function $\hat{f}(\mathbf{x}, h_m, w_m)$ in \cref{eq:2d_rope_decoupled} as follows:
\begin{equation}
\begin{aligned}
\hat{f}(\mathbf{x}, h_m, w_m) &=  \left[ 
(x_0 + ix_1)e^{ih_m\theta^h_0}, (x_2 + ix_3)e^{ih_m\theta^h_1}, \ldots, (x_{d/2-2} + ix_{d/2-1})e^{ih_m\theta^h_{d/4-1}}, 
\right. \\
& \quad  \left.
(x_{d/2} + ix_{d/2+1})e^{iw_m\theta^w_0}, (x_{d/2+2} + ix_{d/2+3})e^{iw_m\theta^w_1}, \ldots, (x_{d-2} + ix_{d-1})e^{iw_m\theta^w_{d/4-1}} 
\right]^T .
\end{aligned} 
\end{equation}

\section{Limitations and Future Work}
Constrained by limited computational resources, we only train FiT-XL/2 for 1800K steps. At the resolution of 256x256, the performance of FiT-XL/2 is slightly inferior compared to the DiT-XL/2 model. 
On the other hand, we have not yet thoroughly explored the generative capabilities of the FiT-XL/2 model when training with higher resolutions (larger token length limitation).
Additionally, we only explore resolution extrapolation techniques that are training-free, without delving into other resolution extrapolation methods that require additional training.
We believe that FiT will enable a range of interesting studies that have been infeasible before and encourage more attention towards generating images with arbitrary resolutions and aspect ratios.

\section{More Model Samples}
\label{appendix_sample}

\begin{figure}[t]
    \centering
    \includegraphics[width=1.0\linewidth]{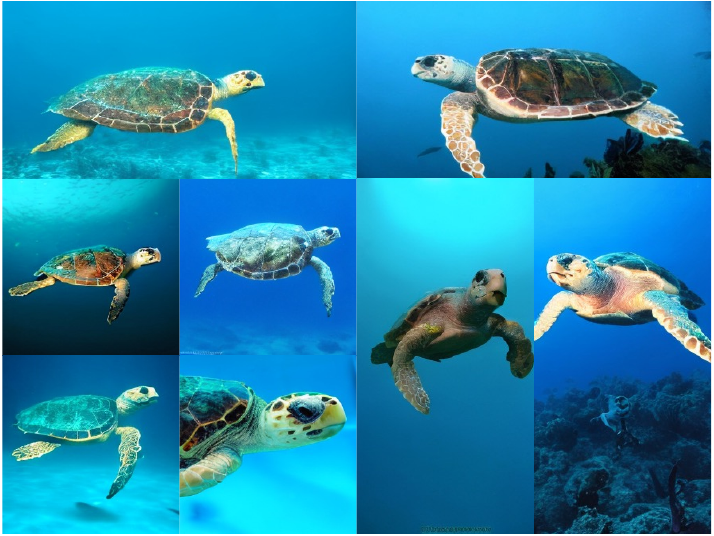}
    \caption{
        Uncurated samples from FiT-XL/2 models at resolutions of $256\times256$, $224\times448$ and $448\times224$.
    }
    \label{fig:sample1}
\end{figure}

\begin{figure}[t]
    \centering
    \includegraphics[width=1.0\linewidth]{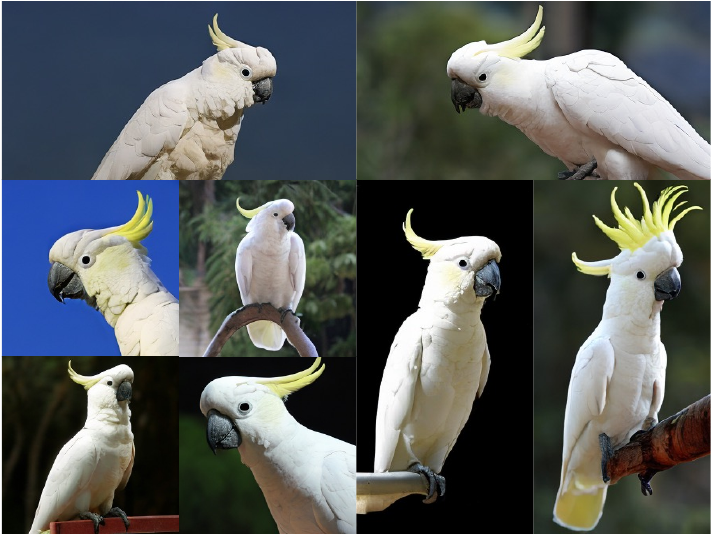}
    \caption{
        Uncurated samples from FiT-XL/2 models at resolutions of $256\times256$, $224\times448$ and $448\times224$.
    }
    \label{fig:sample2}
\end{figure}

\begin{figure}[t]
    \centering
    \includegraphics[width=1.0\linewidth]{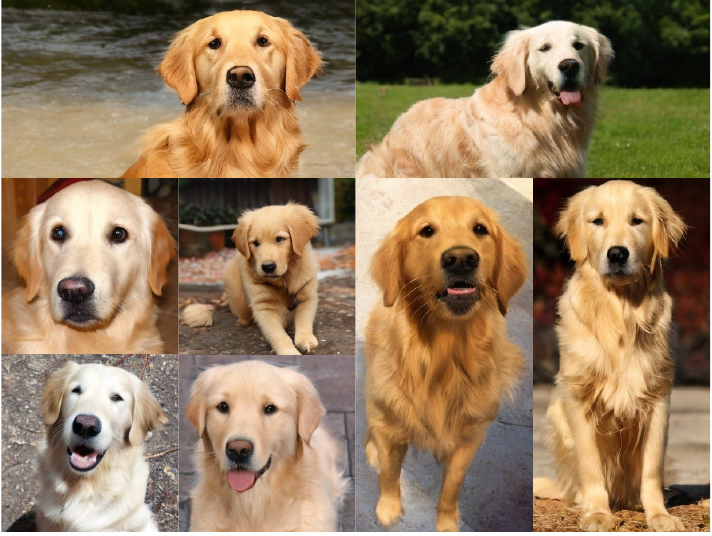}
    \caption{
        Uncurated samples from FiT-XL/2 models at resolutions of $256\times256$, $224\times448$ and $448\times224$.
    }
    \label{fig:sample3}
\end{figure}

\begin{figure}[t]
    \centering
    \includegraphics[width=1.0\linewidth]{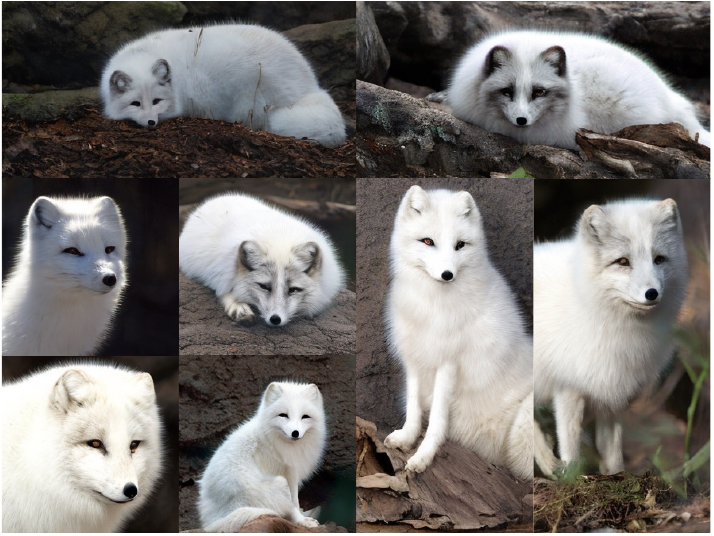}
    \caption{
        Uncurated samples from FiT-XL/2 models at resolutions of $256\times256$, $224\times448$ and $448\times224$.
    }
    \label{fig:sample4}
\end{figure}

\begin{figure}[t]
    \centering
    \includegraphics[width=1.0\linewidth]{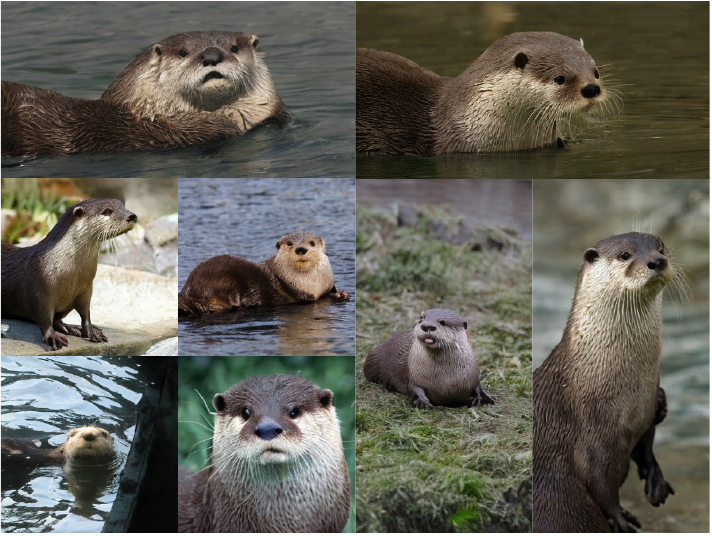}
    \caption{
        Uncurated samples from FiT-XL/2 models at resolutions of $256\times256$, $224\times448$ and $448\times224$.
    }
    \label{fig:sample5}
\end{figure}

\begin{figure}[t]
    \centering
    \includegraphics[width=1.0\linewidth]{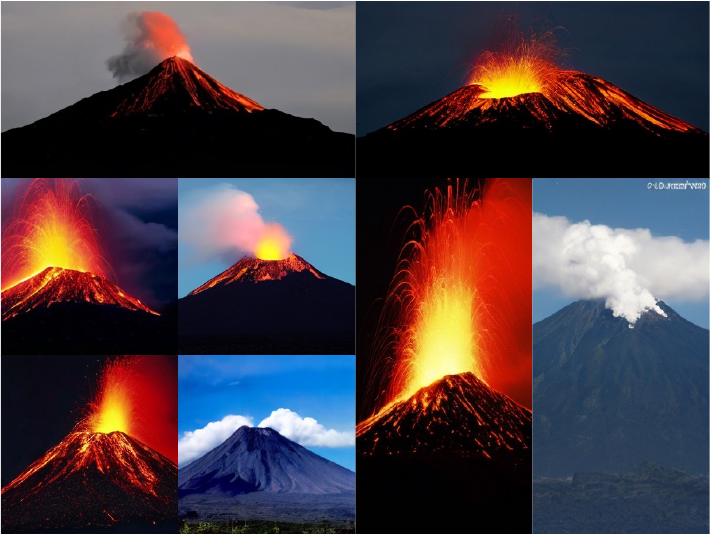}
    \caption{
        Uncurated samples from FiT-XL/2 models at resolutions of $256\times256$, $224\times448$ and $448\times224$.
    }
    \label{fig:sample6}
\end{figure}

We show samples from our FiT-XL/2 models at resolutions of $256\times256$, $224\times448$ and $448\times224$, trained for 1.8M (generated with
250 DDPM sampling steps and the ft-EMA VAE decoder). 
Fig.~\ref{fig:sample1} shows uncurated samples from FiT-XL/2 with classifier-free guidance scale 4.0 and class label ``loggerhead turtle'' (33).
Fig.~\ref{fig:sample2} shows uncurated samples from FiT-XL/2 with classifier-free guidance scale 4.0 and class label ``Cacatua galerita'' (89).
Fig.~\ref{fig:sample3} shows uncurated samples from FiT-XL/2 with classifier-free guidance scale 4.0 and class label ``golden retriever'' (207).
Fig.~\ref{fig:sample4} shows uncurated samples from FiT-XL/2 with classifier-free guidance scale 4.0 and class label ``white fox'' (279).
Fig.~\ref{fig:sample5} shows uncurated samples from FiT-XL/2 with classifier-free guidance scale 4.0 and class label ``otter'' (360).
Fig.~\ref{fig:sample6} shows uncurated samples from FiT-XL/2 with classifier-free guidance scale 4.0 and class label ``volcano'' (980).

\begin{figure}[t]
    \centering
    \includegraphics[width=1\linewidth]{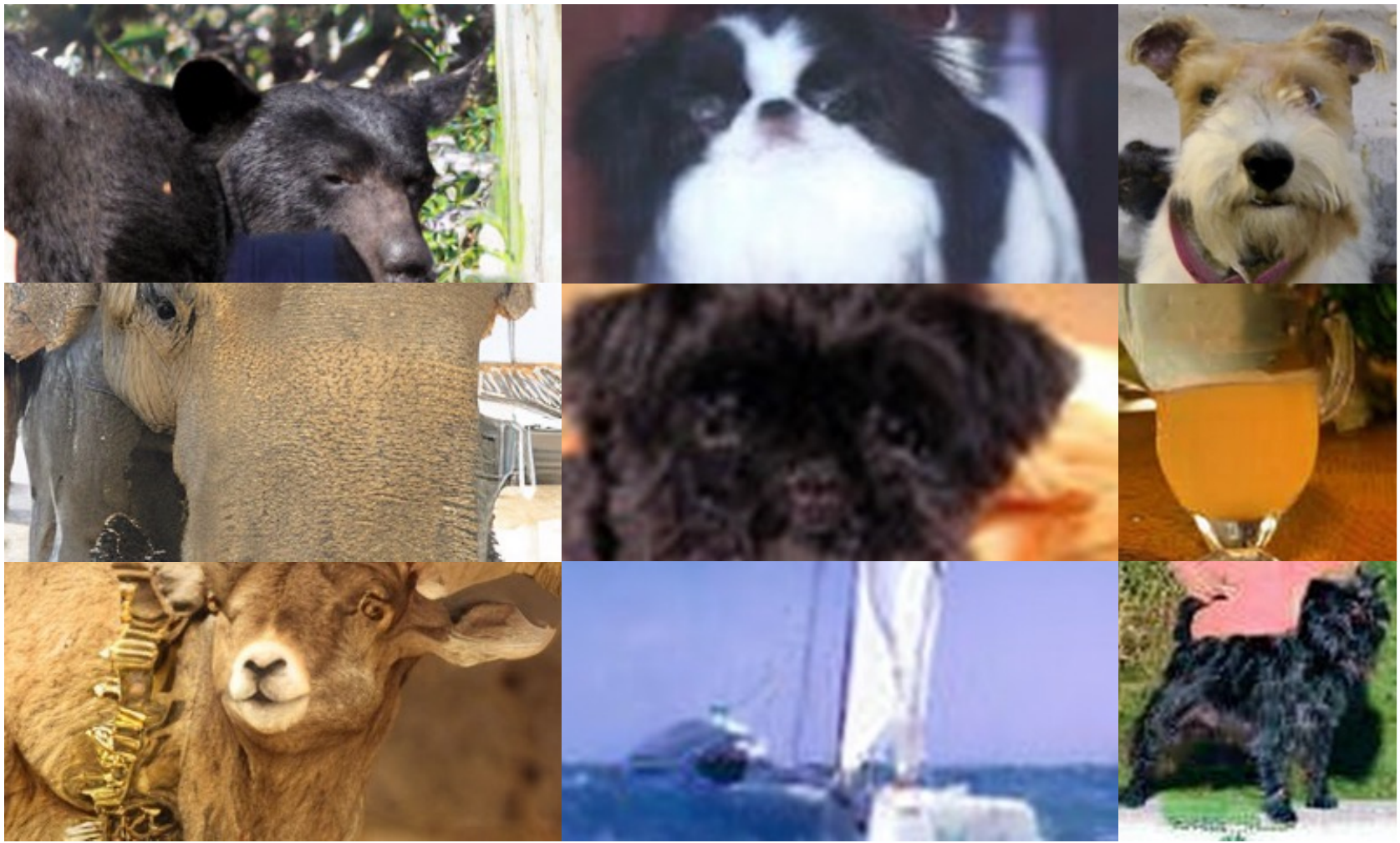}
    \vspace{-0.3cm}
    \caption{
        Uncurated failure samples from DiT-XL/2.
    }
    \vspace{-0.7cm}
    \label{fig:ditbad}
\end{figure}

We also show some failure samples from DiT-XL/2, as shown in Fig.~\ref{fig:ditbad}.

\begin{figure}[t]
    \centering
    \includegraphics[width=0.45\linewidth]{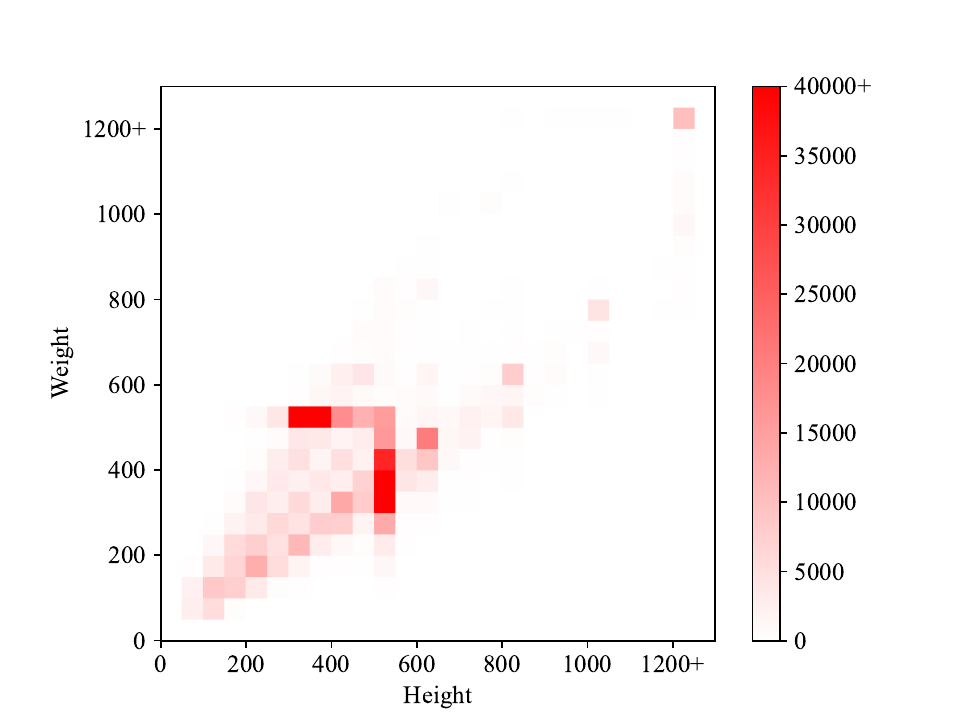}
    \vspace{-0.3cm}
    \caption{
        Height/Width distribution of the original \textit{ImageNet}~\cite{deng2009imagenet} dataset.
    }
    \vspace{-0.7cm}
    \label{fig:imagenet}
\end{figure}

\twocolumn

%%%%%%%%%%%%%%%%%%%%%%%%%%%%%%%%%%%%%%%%%%%%%%%%%%%%%%%%%%%%%%%%%%%%%%%%%%%%%%%
%%%%%%%%%%%%%%%%%%%%%%%%%%%%%%%%%%%%%%%%%%%%%%%%%%%%%%%%%%%%%%%%%%%%%%%%%%%%%%%

\end{document}